\documentclass{article} 
\usepackage{graphicx}
\usepackage{float}
\usepackage{iclr2027_conference,times}

\usepackage{amsmath,amsfonts,bm}

\def\eqref#1{equation~\ref{#1}}

\def\1{\bm{1}}

\DeclareMathAlphabet{\mathsfit}{\encodingdefault}{\sfdefault}{m}{sl}
\SetMathAlphabet{\mathsfit}{bold}{\encodingdefault}{\sfdefault}{bx}{n}

\usepackage{hyperref}
\usepackage{url}
\usepackage{booktabs}
\usepackage{multirow}
\usepackage{amsmath,amssymb}
\usepackage{xcolor}

\title{Video2SwimFish: An Automated Pipeline for Reconstructing Controllable Fish Models and Biological Locomotion from Real Fish Videos}

\author{
Hangong Chen$^{1}$, Linfeng Cheng$^{2}$, Tahsin Zaman Jilan$^{1}$, \\
\textbf{Ian Fuller\textsuperscript{2}, Jiaye Wu\textsuperscript{2}, Lee Caesar\textsuperscript{1}, Yantian Zha\textsuperscript{1*}} \\
$^{1}$Department of Computer Science, North Carolina A\&T State University \\
$^{2}$Department of Computer Science, University of Maryland, College Park \\
$^{*}$Corresponding author: \texttt{yzha@ncat.edu}
}

\iclrfinalcopy 
\begin{document}

\makeatletter
\let\@oldmaketitle\@maketitle
\renewcommand{\@maketitle}{%
  \@oldmaketitle
    \vspace*{-10mm}
  \begin{minipage}{\linewidth}
    \setcounter{figure}{0}
    \refstepcounter{figure}
    \includegraphics[width=\linewidth]{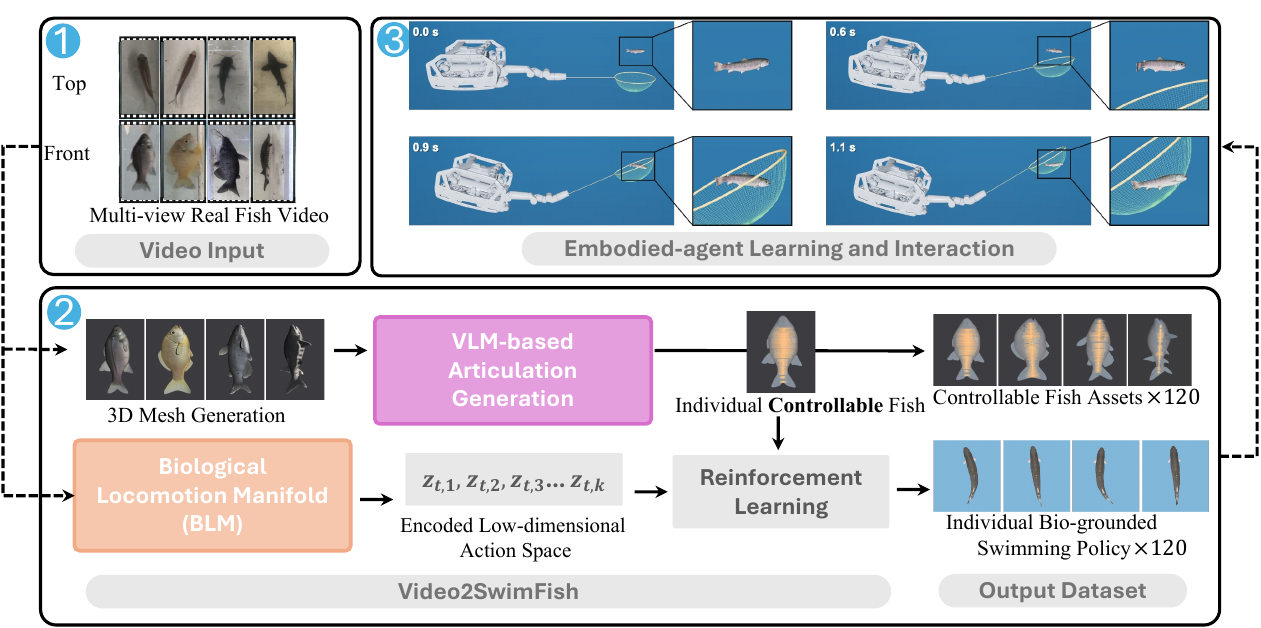}%
    \par\smallskip
    Figure 1: Overview of Video2SwimFish. Given multi-view real fish videos, Video2SwimFish generates individual fish meshes with controllable articulations and derives a Biological Locomotion Manifold (BLM) from real-fish swimming motion. Reinforcement learning then uses the BLM to learn an individual swimming policy for each controllable fish, producing a dataset of biologically grounded, controllable fish assets for downstream embodied-agent learning and interaction.
    \label{fig:overview} 
  \end{minipage}
  \bigskip
  \setcounter{figure}{1}\vspace*{-2mm} 
}
\makeatother

\maketitle
\lhead{Preprint}

\begin{abstract}
We present \textbf{Video2SwimFish}, an automated pipeline and benchmark for building controllable fish assets from real-fish videos for underwater embodied AI. Given synchronized multi-view videos of an individual fish, the pipeline generates a metrically scaled deformable mesh from a VLM-selected canonical frame, builds internal articulation adapted to that individual's morphology through a VLM actor--critic loop, and extracts a Biological Locomotion Manifold (BLM). The BLM provides a low-dimensional action space bounded by real-fish motion. Compared with learning directly in joint space, it lets the individual swimming policy of each reconstructed fish converge faster and reach a higher task success rate. We release two paired datasets: synchronized top- and front-view recordings of 120 individual fish across 6 species, and the controllable assets and individual swimming policies derived from them. Because every asset is tied to the animal it came from, the dataset supports a benchmark that evaluates locomotion learning not only on task success but on fidelity to that individual's locomotion behavior. We evaluate on two tasks, 3D trajectory following and target reaching, and a case study in which a simulated BlueROV underwater robot captures the assets. We further validate biological realism with a human study in which fish experts rate the simulated fish, as a complement to the automated metrics. We find that task success and locomotion fidelity do not necessarily improve together, and we identify faithful reproduction of individual animal locomotion as an open challenge for the community. Project website: \url{https://hangongchen.github.io/video2swimfish-web/}.
\end{abstract}
\section{Introduction}

Embodied agents trained with reinforcement learning can monitor, inspect, and manage
aquaculture environments
\citep{fore2018precision,amundsen2024aquaculture,kelasidi2025autonomous,sumana2026embodied},
which lowers labor costs, keeps workers out of hazardous sea-cage operations, and
supports sustainable seafood production for global food security
\citep{fao2024sofia,fao2025guidelines}. However, training
embodied agents in real-world aquaculture environments is costly, hard to scale
\citep{amundsen2024aquaculture}, and involves potential risks to animal welfare
\citep{pino2024towards}. Simulation provides a safer and more scalable alternative. Existing
simulation frameworks can already model many of the physical components needed for underwater
robot learning, including fluid systems \citep{todorov2012mujoco}, articulated robot dynamics
\citep{makoviychuk2021isaac}, and deformable-body dynamics
\citep{faure2012sofa,hu2019chainqueen}. A major gap, however, is \textit{the lack of realistic,
deformable, and controllable fish with a swimming policy that are both grounded in real-world
observations, together with standardized tools for
evaluating locomotion and biological fidelity in simulation}. Real-fish videos provide a low-cost, non-invasive, and scalable source of
biological observations that can capture both fish morphology and swimming behavior~\citep{cui2025fish}. Thus, we
present \textbf{Video2SwimFish}, an end-to-end pipeline that reconstructs fish models and learns
biological locomotion from real fish videos. The pipeline can be reused by users without specialized biological knowledge to generate
bio-grounded, controllable fish from new real-fish videos, each equipped with an individual
swimming policy.

Building such a fish from real-world videos presents two main challenges. Reconstructing an articulated 3D fish from video is the first challenge. NeRF-based methods are
effective for reconstructing 3D shapes and deformable objects from images and videos
\citep{mildenhall2020nerf,pumarola2021dnerf,yang2022banmo}, but they do not directly produce the
internal articulation needed to control a simulated fish and perform poorly when reconstructing
fast-moving, highly deformable fish. Recent VLM-based methods such as URDFormer and
Articulate-Anything can further turn visual inputs into articulated simulation assets
\citep{chen2024urdformer,le2024articulate}. However, these methods target rigid articulated objects, whereas a fish is a deformable body whose internal articulation must be inferred and coupled with the mesh. 

The second challenge is to
learn a biologically grounded swimming policy. In order to ensure that the learned policy is
grounded in biological locomotion, we therefore train it to follow the swimming motion observed
in the real-fish video. Existing approaches can learn locomotion using reinforcement learning \citep{schulman2017proximal},
imitation learning~\citep{ross2011dagger,peng2021amp}, or hand-designed locomotion models such as central pattern generators (CPGs)
\citep{xie2019central}. However, CPGs impose a predefined locomotion pattern rather
than being grounded in the motion of each individual fish, while imitation learning typically
requires corresponding joint-level demonstrations. In our setting, the joint states underlying
the motion in a real-fish video are not directly available, making direct joint-level imitation
difficult. Reinforcement learning can learn task-directed locomotion, but task rewards alone do
not specify how the fish should swim. For example, a fish could complete the task yet move very unlike a real fish.

To address these challenges and enable systematic evaluation, our contributions in
Video2SwimFish are fourfold:

\textbf{1. Video2SwimFish Pipeline:} An end-to-end pipeline that turns real fish
videos into realistic, controllable, deformable simulated fish, each with its own swimming policy.

\textbf{2. Biological Locomotion Manifold (BLM):} A low-dimensional action space
extracted from each fish's own 3D swimming motion, so that every policy moves within
the real animal's range of bending. Compared with joint-space RL, BLM enables a higher
completion rate and covers more of the real fish's motion.

\textbf{3. Datasets:} To our knowledge\footnote{The closest prior work, 3D-ZeF \citep{pedersen20203d}, records in total 32 fish of one species across its 8 sequences. Rather than individual morphology, it outputs only movement trajectories. Ours contain exactly
one identified fish each, across 120 individuals and 6 species.}, the largest
synchronized multi-view fish video dataset: 120 individual fish across 6 species,
4$\times$ more than existing fish datasets. Running our Video2SwimFish
pipeline on these videos produces 120 simulated fish, each with its own swimming
policy. They are realistic, moving and interactable, and can be used by the community
to train and test underwater robots.

\textbf{4. Benchmark:} Our benchmark evaluates locomotion learning not only on task
completion but also on biological fidelity, measured by trajectory shape, body
bending and swimming speed, together with an expert human study, showing that BLM+RL outperforms existing baselines in overall task success and core biological fidelity metrics. A BlueROV2 capture case study further shows that the
fish serve real underwater robotics needs.

\section{Related Work}
\label{sec:related}

\textbf{From Video to Controllable 3D Assets.}
NeRF \citep{mildenhall2020nerf} and its dynamic extensions model deformable
objects with canonical representations and deformation fields
\citep{pumarola2021dnerf,park2021nerfies}, and BANMo recovers animatable 3D
models from casual videos \citep{yang2022banmo}; none provides the internal
articulation needed for control, and all struggle with fast, highly deformable
motion. Single-image reconstruction methods
\citep{hong2024lrm,tochilkin2024triposr,long2024wonder3d,xu2024instantmesh}
generate 3D assets efficiently, while URDFormer and Articulate-Anything build
articulated simulation assets from visual inputs
\citep{chen2024urdformer,le2024articulate}, but target static objects or rigid
articulations. Skinned NeRF or BANMo meshes can deform but have no physics,
and simulation-ready methods assume rigid parts. \emph{Our novelty} is a hybrid
asset that is both deformable and controllable: a rigid articulation placed
inside a soft FEM body, enabling physics simulation and RL of a deforming fish.

\textbf{Learning Locomotion from  Videos.}
Imitation-from-observation methods learn behaviors from state or video
demonstrations without action labels
\citep{liu2017imitation,torabi2018bco,edwards2019ilpo,raychaudhuri2021crossdomain},
but rely on an inverse dynamics model whose pseudo-labels compound error when
simulated and real dynamics diverge. Reference-motion methods such as DeepMimic and AMP
\citep{peng2018deepmimic,peng2021amp} instead combine reference motions with
reinforcement learning, but need joint-level references that real-fish videos do
not provide. For fish-like robots, CPGs and RL over their parameters generate
rhythmic swimming \citep{ijspeert2008central,wang2012multimodal,
zhang2020path,deng2024robot,rodwell2023physics}, yet those
parameters are shared across individuals. To address this individual-specific locomotion learning problem,
we propose the BLM. \emph{Its unique value} is to accurately extract each fish's own
body motion from video and encode it as the low-dimensional action space of its
policy, so that each digital fish moves more like its real-world counterpart
while training faster.

\textbf{Simulation Assets and Embodied-Learning Benchmarks.}
Large-scale 3D asset datasets \citep{deitke2023objaverse,wu2023omniobject3d} and
articulated-object datasets \citep{mo2019partnet} supply models for
simulation, and platforms such as SAPIEN, MuJoCo, Isaac Gym and Isaac Sim
simulate articulated and deformable bodies at scale
\citep{xie2020sapien,todorov2012mujoco,makoviychuk2021isaac,mittal2023isaacsim}.
On these, benchmarks such as AI2-THOR, Habitat, ManiSkill and BEHAVIOR-1K
standardize embodied-agent training
\citep{xia2020interactive,szot2021habitat,gu2023maniskill2,li2024behavior1k}, and
RoboGen generates tasks and scenes with foundation models
\citep{wang2024robogen}. For underwater robotics, FishGym provides articulated
fish, fluid--structure interaction and RL \citep{liu2022fishgym}, but its fish
are hand-designed rather than grounded in biological observation, and no
standardized evaluation of biological fidelity exists. \emph{The practical benefit} of our pipeline
is the scale and scalability of its dataset: 120 automatically built fish of 6 species,
each with its own policy and benchmark metrics, enabling future applications as moving
targets in underwater robot learning and biological locomotion learning evaluation against
each real fish.

Table~\ref{tab:novelty} (Appendix~\ref{app:novelty}) compares our novelty with prior work item by item.

\section{Video2SwimFish}

\textbf{Problem Statement.}
Given a set of real-fish videos $\mathcal{V}=\{V_i\}_{i=1}^{N}$, each a
synchronized top- and front-view recording of one individual, our goal is to
produce for each a simulated fish $F_i=(M_i,K_i,\pi_i)$ that can autonomously
swim in a physics-based environment: a deformable mesh $M_i$ preserving that
fish's morphology, an internal articulation
$K_i=(\mathcal{B}_i,\mathcal{J}_i,\Psi_i)$ of rigid bones $\mathcal{B}_i$,
joints $\mathcal{J}_i$ with their limits and drive parameters, and an
attachment $\Psi_i$ binding mesh to bones so that actuating $\mathcal{J}_i$
deforms $M_i$, and a swimming policy $\pi_i$ consistent with the motion
observed in $V_i$. Unlike rigid articulated objects, the bones are internal and
the observable body is deformable. Ground-truth kinematics are not available
from video, so $K_i$ cannot be optimized directly; its quality is instead
reflected in $\pi_i$, since a poorly constructed articulation limits the
reachable motion space. 

\begin{figure}[H]
    \centering
    \includegraphics[width=1.0\linewidth]{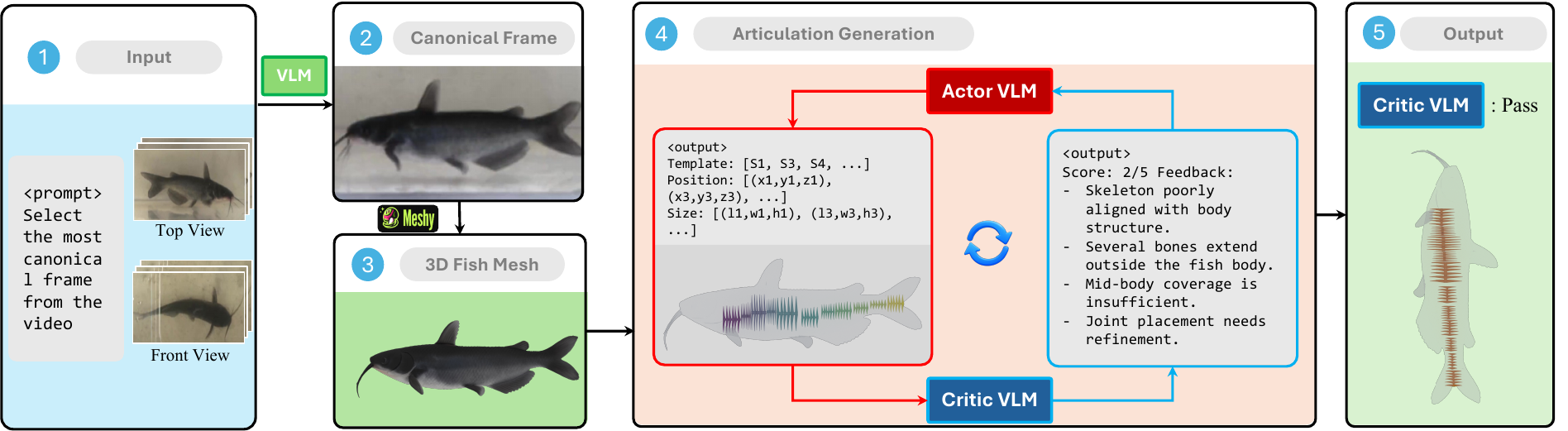}
\caption{The Video2SwimFish articulation generation pipeline. A VLM selects the most canonical frame from the multi-view video, from which Meshy generates a 3D fish mesh. An actor VLM then proposes a skeleton as a list of bone templates with positions and sizes, and a critic VLM scores it and returns feedback; the two iterate until the critic accepts.}
    \label{fig:method-overview}
\end{figure}

Our framework generates $M_i$ and $K_i$ from the
video (Figure~\ref{fig:method-overview}), extracts the observed swimming motion into a low-dimensional Biological
Locomotion Manifold (BLM), and uses that manifold as the action space for
goal-conditioned reinforcement learning, which trains $\pi_i$. Applied across
$\mathcal{V}$, this yields a collection of simulated fish, each with its own morphology, articulation, and swimming policy.

\subsection{Mesh Generation}

\label{sec:mesh-reconstruction}

A fish video captures continuously changing body poses, and different poses can introduce transient body deformations that distort the morphology in a single frame. Since image-to-3D generation operates on a single image, selecting an arbitrary frame may therefore produce a mesh that reflects a transient swimming pose rather than a stable morphology. We therefore first identify a canonical\footnote{Canonical here means the pose closest to the fish's resting
morphology, a standard notion in deformable 3D reconstruction
\citep{pumarola2021dnerf,park2021nerfies,yang2022banmo}.} body-pose frame from the video.

For this step, we use a pretrained Qwen3-VL model 
\citep{qwen3vl2025} to evaluate the body configuration 
of each candidate frame and select the frame that best 
represents the fish's canonical morphology. The selected frame is cropped to the fish region using 
background subtraction \citep{zivkovic2004background} 
against the per-video median background, so that the 
image-to-3D system receives the fish rather than the 
tank environment. The selected frame is then used as the visual reference for subsequent 3D mesh generation.

The selected canonical frame is then given to Meshy, an off-the-shelf image-to-3D system, to
generate the three-dimensional fish mesh \citep{meshy_image_to_3d}\footnote{We explored a range of video- and image-based reconstruction
methods, including deformable NeRF variants
\citep{pumarola2021dnerf,park2021nerfies,yang2022banmo}, feed-forward
image-to-3D models \citep{tochilkin2024triposr,long2024wonder3d,xu2024instantmesh},
and VLM-driven voxel generation. The NeRF-based methods degraded on fast,
non-rigid fish motion, the feed-forward models produced fused or truncated
fins, and voxel generation was limited by context length to resolutions too
coarse for thin structures such as fins. Meshy gave the most usable
meshes.}. To scale the mesh to the fish's actual size, we estimate each fish's body 
length from
video, calibrated using the known tank dimensions 
visible in both views. The generated mesh 
is scaled to this estimated length before the subsequent 
articulation construction stage. This yields a mesh consistent with the morphology of the real fish for articulation construction.

The generated mesh is hard to validate, since 3D ground truth of a live fish is difficult to obtain
\citep{pumarola2021dnerf,park2021nerfies,yang2022banmo}. We therefore evaluate the
meshes in two indirect ways: the silhouette IoU between each mesh and hand-drawn outlines of
the real fish (mean affine-aligned IoU of $0.91$ in the input side view and $0.75$ in a top
view that the image-to-3D model never saw; Appendix~\ref{app:mesh-iou}), and the
\emph{appearance} score of the human study (mean $4.9$/$5$ over all clips; Section~\ref{sec:human-study}).

\subsection{Actor--Critic Auto-Construction of Fish Articulation }

The limitation of a generated fish mesh is that it does not specify how the fish should be controlled. In particular, it does not provide an internal skeleton, joint locations, joint configurations, or the physical parameters required to construct a controllable fish, including joint stiffness, damping, and actuator properties. Designing these by hand for every fish would be expensive and hard to scale.

Following recent VLM-based actor--critic approaches for articulated object modeling \citep{le2024articulate}, we employ a VLM actor--critic framework to automatically construct the internal articulation of each reconstructed fish.\footnote{Our goal is not to recover the anatomical skeleton of each fish. A real fish is driven by muscles acting on bones and soft tissue, while our simulated fish is driven by joint torques under an approximate hydrodynamic model, so even an exact copy of the real skeleton would not by itself swim like the real fish. We therefore only require the articulation to be a plausible and stable control structure on which RL can learn, and we judge biological fidelity by how the resulting fish swims (Section~\ref{sec:benchmark}).} Note that this VLM-based actor--critic is distinct from the reinforcement-learning actor--critic used for policy training \citep{sutton1998reinforcement}. Here, the actor VLM proposes an articulation configuration, while the critic VLM evaluates its physical plausibility and compatibility with the fish morphology and provides feedback to the actor VLM. This actor--critic process is iterated until the critic VLM determines that the configuration reaches our empirically defined satisfactory score.

In our design, particularly, the actor VLM is a pretrained Qwen3-VL model fine-tuned on a small collection of manually constructed fish models. Our manually constructed dataset contains 74 training examples for 
fine-tuning the actor VLM. These examples provide the actor VLM with fish-specific knowledge about how an articulated structure should be placed inside a fish mesh.

Rather than generating an arbitrary skeleton directly, we provide the actor VLM with a library of skeleton templates (Appendix~\ref{app:skeleton-dataset}). A template represents a valid bone segment that can be placed inside the fish body. The actor VLM selects templates and determines their locations and sizes to construct the articulation. At each iteration, the actor VLM observes rendered views of the fish mesh together with the current articulation and proposes a modification to the current configuration. The modification may add, remove, reposition, or resize 
a bone. We adopt a fixed physically validated preset 
for joint DOFs, limits, and drive parameters shared 
across all fish. Since the simulated fish is not required 
to replicate the mechanical constraints of a real fish 
body, this design gives the reinforcement learning policy 
a sufficiently rich action space to discover effective 
swimming gaits. Per-bone mass is derived from the estimated body volume, in proportion to bone volume.

At iteration $k$, with articulation $K^{(k)}$, the actor predicts
\begin{equation}
    a^{(k)} = \mathcal{A}_\phi\left(M, K^{(k)}, f^{(k-1)}\right)
\end{equation}
where $M$ is the generated fish mesh described in 
Section~\ref{sec:mesh-reconstruction}, and $f^{(k-1)}$ 
is the critic VLM feedback from the previous iteration 
(empty at $k=0$). At each iteration, the actor VLM observes 
three rendered views of the fish mesh with the current 
articulation highlighted beneath a semi-transparent skin, 
and receives the measured offset of each bone from the 
body centerline as quantitative context alongside the 
critic VLM feedback. The proposed modification produces a 
new articulation
\begin{equation}
    K^{(k+1)} = \mathcal{T}\left(K^{(k)}, a^{(k)}\right)
\end{equation}

The proposed articulation is then evaluated by the critic VLM. The critic VLM receives the fish mesh and the proposed articulation and provides a score together with feedback describing the problems that should be corrected. The critic VLM feedback can identify issues such as inappropriate bone placement, insufficient coverage of the body, or implausible articulation structure. Formally,
\begin{equation}
    \left(
        r^{(k)},f^{(k)}
    \right)
    =
    \mathcal{C}_{\psi}
    \left(
        M,K^{(k+1)}
    \right)
\end{equation}
where $r^{(k)}$ is the critic score and $f^{(k)}$ its feedback.

The actor VLM uses this feedback to update the articulation and proposes the next modification. The loop runs for a fixed budget of $K$ iterations or 
until the critic VLM accepts. If the actor VLM oscillates between 
adding and removing the same bone, its sampling 
temperature is raised to escape the cycle. Upon 
termination, the highest-scoring articulation across 
all iterations is selected and passed through a 
deterministic structural completion step $\mathcal{R}$, 
which fills internal gaps and extends the chain toward head and tail:
\begin{equation}
    K^* = \mathcal{R}\left(K^{(k)}\right)
\end{equation}

The final articulation $K^{*}$ specifies the bones, joints, and physical parameters of the fish. The deformable fish mesh is then coupled to this internal articulated structure in the physics simulator. The resulting fish is therefore both deformable and controllable, with its individual morphology determined by the generated mesh and its control structure determined by the articulation auto-construction process.

\subsection{Biological Locomotion Manifold (BLM)}

Real-fish videos capture swimming as continuous body deformation rather than explicit joint-level motion. In contrast, the reconstructed fish is actuated through its internal joints, making it difficult to directly learn the locomotion from real-fish videos. We therefore introduce a \textit{Biological Locomotion Manifold (BLM)}, a low-dimensional representation of three-dimensional biological body deformation extracted from synchronized top and side-view of individual fish, expressed directly in the joint space of the reconstructed fish, which enables reinforcement learning to control the fish through biologically grounded motion.

We represent fish body deformation by the three-dimensional shape of the body midline,
recovered from two synchronized views. In each frame we extract the 2D midline from the
top view and from the side view and fuse them into a 3D midline: the longitudinal and
lateral coordinates come from the top view, and the vertical coordinate from the side
view. The midline is expressed in a head-anchored body frame and normalized by body
length (BL).

Instead of describing this midline by curvature, we express it directly in the joint
space of the reconstructed fish. Let the fish have $N$ joints along the body. At timestep
$t$, we fit its bone chain to the 3D midline by exact sequential inverse kinematics: each
bone is aligned with the midline tangent at its position, and the relative rotation
between consecutive bones is split into a lateral (yaw) angle $\psi$ and a vertical
(pitch) angle $\theta$. The body pose is
\begin{equation}
    \mathbf{c}_t
    =
    \left[
        \psi_1(t),\ldots,\psi_N(t),\;
        \theta_1(t),\ldots,\theta_N(t)
    \right]^\top
    \in \mathbb{R}^{2N}.
\end{equation}
Roll about the body axis cannot be observed from the two silhouettes and is set to zero.
Because the fit uses the fish's own bone layout, $\mathbf{c}_t$ can be realized directly
by its joints.

We collect these joint-angle profiles from real-fish videos and apply principal component
analysis (PCA) \citep{pearson1901liii} to extract a low-dimensional biological action
representation. Since lateral and vertical angles differ in magnitude, each half is
normalized by its own standard deviation,
$\Sigma=\operatorname{diag}(\sigma_\psi\mathbf{1}_N,\sigma_\theta\mathbf{1}_N)$.
Let $\boldsymbol{\mu}$ denote the mean pose and $V\in\mathbb{R}^{2N\times d}$ the first $d$
principal components. A body pose can then be approximated as
\begin{equation}
    \mathbf{c}_t \approx \boldsymbol{\mu} + \Sigma V\mathbf{a}_t,
\end{equation}
where $\mathbf{a}_t\in\mathbb{R}^{d}$ contains the PCA coefficients. Because a single PCA
covers both halves, each mode couples lateral and vertical bending, as in real fish.

We use the PCA coefficients as the biological action representation. In our main
experiments, we retain the first 8 principal components, which capture 99.6\% of the
joint-angle variance on average; truncating to these 8 modes moves the 3D midline by only
0.001\,BL. The remaining modes add almost no expressive range but would contribute
action noise in every exploration step. The range and temporal variation of this action
representation are derived directly from the real-fish trajectories. Since the video
frame rate $f_v$ differs from the control rate $f_c$, for each PCA coefficient we compute
its absolute change per control step
\begin{equation}
    \delta a_{i,t} = \frac{f_v}{f_c}\left|a_{i,t+1}-a_{i,t}\right|
\end{equation}
and define the maximum per-step change as the 99th percentile over the real-fish data:
\begin{equation}
    \Delta a_{i,\max} = P_{99}\left(\delta a_{i,t}\right).
\end{equation}
The resulting vector
\begin{equation}
    \Delta\mathbf{a}_{\max} = \left[\Delta a_{1,\max},\ldots,\Delta a_{d,\max}\right]^\top
\end{equation}
therefore defines a biologically grounded bound on how rapidly each PCA coefficient can
change. The coefficient limits are likewise determined from the observed real-fish
trajectories, $a_{i,\max}=P_{99}\left(|a_{i,t}|\right)$.

To check that the simulated fish can actually perform BLM actions, we replay each real fish's
own BLM coefficients open-loop on its asset, for all twelve benchmark fish. The joints follow
the commanded angles with $R^2=0.99$, an amplitude ratio of $0.98$ and no lag at the 33\,ms
control resolution, and the resulting 3D midline stays within $0.007$\,BL of the real one
(Appendix~\ref{app:blm-tracking}).

\subsection{Data Collection}
\label{sec:data-collection}

\textbf{Multi-View Fish Videos.}
We record each fish alone in a transparent tank with two hardware-synchronized cameras, one
top-down and one front-facing, at 32\,fps and $2736 \times 1824$ resolution for two and a half
minutes. The dataset comprises 120 individual fish of 6 species; the recording setup and
per-species details are given in Appendix~\ref{app:dataset}. On the human-annotated subset, the
automatically extracted midlines lie within 0.9\% (top view) and 5.0\% (front view) of body
length of the human ones (median Chamfer distance; human-to-human 0.7\% and 1.6\%), and their
length differs by 9.3\% and 16.4\% (Appendix~\ref{app:midline-acc}).

\textbf{Fish Skeleton Dataset.}
To fine-tune the actor VLM, we manually build fourteen controllable fish models with 85 bones in
total; twelve yield 74 training examples and two are held out for validation. Following
biological references on fish skeletal anatomy, joint mechanics and body stiffness
\citep{lauder2015fish,videler1993fish,long1996importance,jimenez2023flexibility}, we place bone
templates along the body midline of each mesh and assign joint and physical parameters
(Appendix~\ref{app:skeleton-dataset}).

\section{Benchmark}
\label{sec:benchmark}
\label{sec:experiments}

A simulated fish is useful when it both completes its assigned task \emph{and} moves like the
animal it was built from. We therefore evaluate on two swimming tasks, each scored on both task
success and biological fidelity, followed by a downstream robot case study and an expert human
study. We show the key results below.

\textbf{Setup.}\label{sec:setup}
We compare BLM+RL with four baselines: Joint RL (PPO \citep{schulman2017proximal} on joint
targets), Joint RL+AMP \citep{peng2021amp}, CPG+RL \citep{xie2019central} and BCO+RL
\citep{torabi2018bco}. All methods share the same set of individual fish, physics, reward and training settings,
and differ only in the action space or motion prior. Training every method with three seeds on all 120 fish is beyond our compute. We therefore evaluate on a fixed
subset of twelve fish (two randomly selected per species, 10\%) that covers all six species (Appendix~\ref{app:protocol}), which is
common practice for large benchmarks, including BEHAVIOR-1K \citep{li2024behavior1k}, CausalWorld
\citep{ahmed2021causalworld} and RLBench \citep{james2020rlbench,shridhar2022peract}.
The runs of all three random seeds are evaluated on held-out episodes; we report the mean $\pm$
standard deviation over fish. The detailed setup is in Appendix~\ref{app:tasks}.

\subsection{Task 1: Trajectory Following}
\label{sec:task1}

The most direct test of a simulated fish built from video is whether it can swim the same paths
as its real counterpart. In this task, each fish follows held-out 3\,s segments of the 3D
trajectory recorded from its real counterpart, which we call the \emph{reference path}.
We score task success by \emph{completion rate} and \emph{Fr\'echet distance} to the reference
path,\footnote{Reported in body lengths (BL), the fish's snout-to-tail-base length, so that
fish of different sizes can be compared \citep{videler1993fish}.} and biological fidelity by
\emph{motion coverage} (how much of the real fish's body-bending motion the policy
reproduces), \emph{envelope correlation} (how well the bending amplitude along the body
matches) and \emph{speed error} (Appendix~\ref{app:task1}).

Table~\ref{tab:task1} shows that BLM+RL has the best mean on four of the five columns:
it completes 0.70 of the held-out paths, covers 0.85 of the real fish's bending motions
against at most 0.78 for any baseline, and has the bending envelope closest to the real
animal. BLM+RL also learns faster: its median training run reaches a completion rate of 0.5 in 20
epochs, against 32 for BCO+RL and 70 for Joint RL, while the median Joint RL+AMP run never
reaches it.

\begin{table}[H]
\caption{Task~1 3D trajectory following on twelve fish (two per species), evaluated on held-out
windows: completion rate, Fr\'echet distance (BL), motion coverage, bending-envelope
correlation and speed error $\Delta v$ (BL/s). Mean $\pm$ standard deviation over fish; each
fish is first averaged over three seeds. Best mean per column in bold.}
\label{tab:task1}
\centering
\small
\begin{tabular}{@{}lccccc@{}}
\toprule
Method & \shortstack{Completion\\rate\,$\uparrow$} & \shortstack{Fr\'echet\\distance\,$\downarrow$} & \shortstack{Motion\\coverage\,$\uparrow$} & \shortstack{Envelope\\correlation\,$\uparrow$} & \shortstack{Speed\\error $\Delta v$\,$\downarrow$} \\
\midrule
Joint RL & 0.50$\pm$0.29 & 1.14$\pm$0.60 & 0.63$\pm$0.19 & 0.40$\pm$0.16 & 0.23$\pm$0.18 \\
Joint RL + AMP & 0.46$\pm$0.28 & 1.24$\pm$0.66 & 0.61$\pm$0.19 & 0.53$\pm$0.17 & 0.24$\pm$0.18 \\
CPG + RL & 0.68$\pm$0.29 & 1.04$\pm$0.57 & 0.74$\pm$0.20 & 0.46$\pm$0.19 & \textbf{0.17}$\pm$0.07 \\
BCO + RL & 0.52$\pm$0.29 & 1.13$\pm$0.55 & 0.78$\pm$0.18 & 0.44$\pm$0.18 & 0.21$\pm$0.18 \\
\textbf{BLM + RL (Ours)} & \textbf{0.70}$\pm$0.28 & \textbf{0.87}$\pm$0.46 & \textbf{0.85}$\pm$0.16 & \textbf{0.56}$\pm$0.14 & 0.18$\pm$0.14 \\
\bottomrule
\end{tabular}
\end{table}

\begin{figure}[H]
    \centering
    \includegraphics[width=0.95\linewidth]{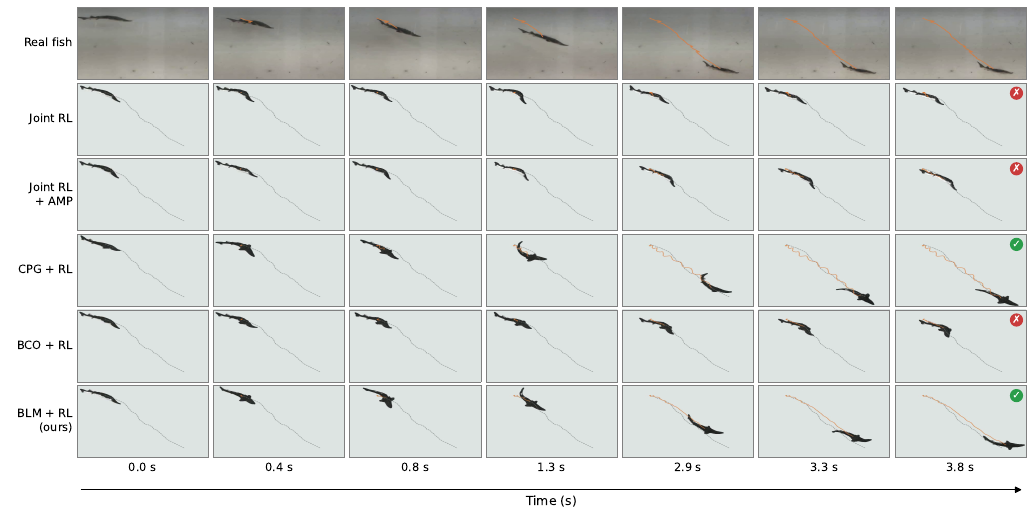}
\caption{The five methods on the same held-out Task~1 trajectory of lake sturgeon 001, side view. Top row: the real fish; other rows: one method each; columns: time. Dashed gray: the real fish's reference trajectory; orange: path swum so far; check/cross: completed or not.}
    \label{fig:fig3}
\end{figure}

\begin{figure}[H]
    \centering
    \includegraphics[width=0.62\linewidth]{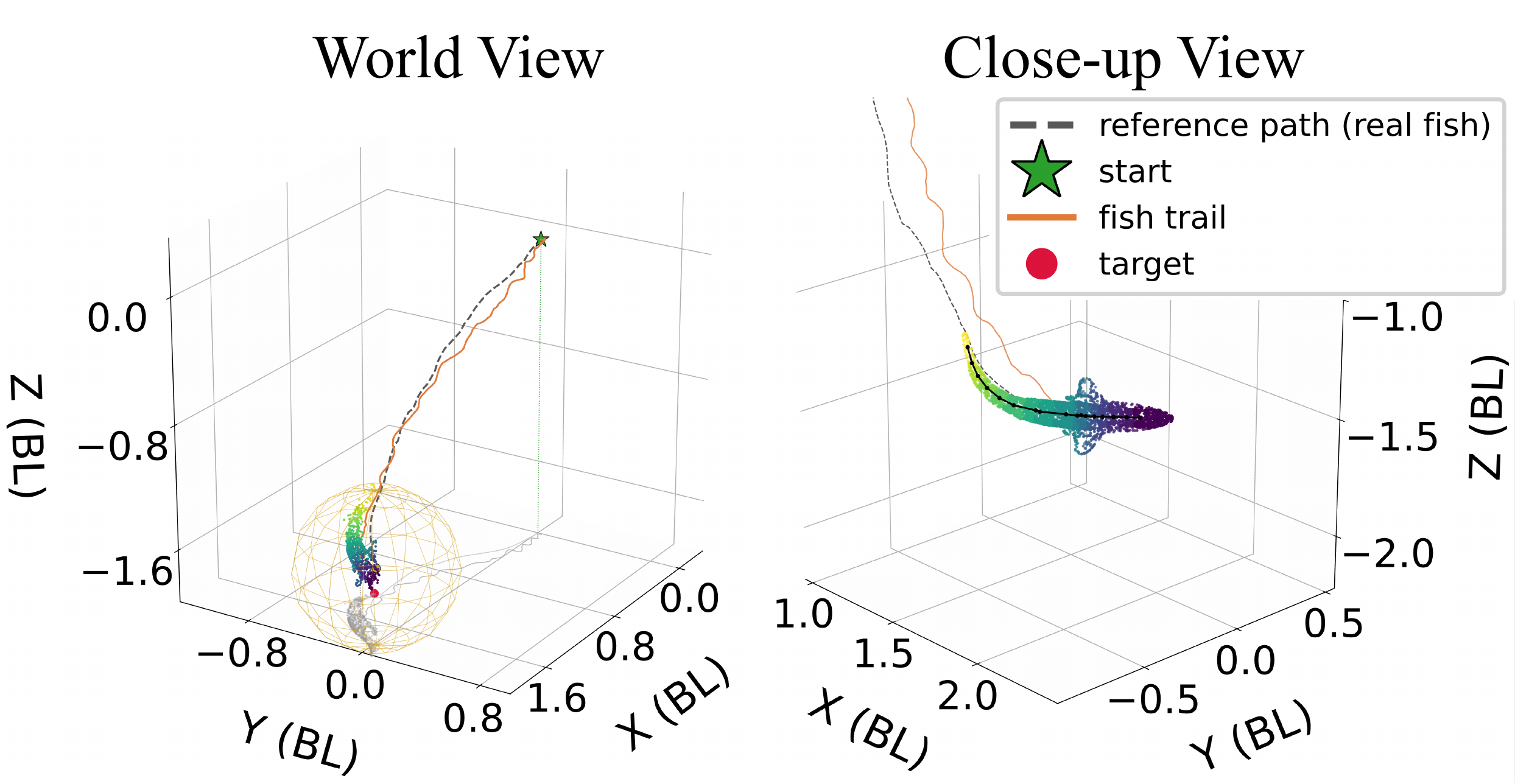}
\caption{The BLM+RL episode of Figure~\ref{fig:fig3} in 3D. Dashed gray: the real fish's reference path; orange: the simulated fish's trail; star: start; red dot: target. World view: the whole episode. Close-up view: the fish's body and midline. Full videos of all methods: \url{https://5-baseline-with-one-real-fish.vercel.app/}.}
    \label{fig:blm3d}
\end{figure}

Figure~\ref{fig:fig3} shows the five methods on a lake sturgeon trajectory that descends while
turning. The three joint-space methods (Joint RL, Joint RL+AMP and BCO+RL) barely leave the
start, while CPG+RL and BLM+RL complete the path. Figure~\ref{fig:blm3d} shows the same
BLM+RL episode in 3D: the fish moves through 3D space in a controlled way, descending and
turning along the reference path.

\begin{figure}[H]
    \centering
    \includegraphics[width=0.60\linewidth]{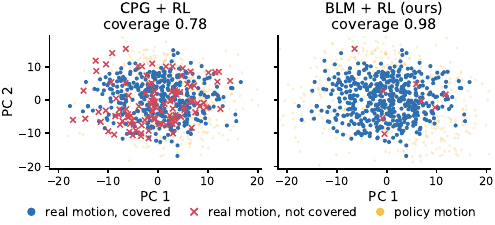}
\caption{Motion coverage on the fish of Figure~\ref{fig:fig3}. Each point is one 0.5\,s body-bending motion of the real fish, projected onto its first two principal components. CPG+RL leaves more of the real motions uncovered than BLM+RL.}
    \label{fig:coverage}
\end{figure}

We attribute these gains to the action space: every BLM action is a coordinated whole-body
bend that the real fish has actually made, so exploration starts from fish-like motion.
CPG+RL is also low-dimensional and completes a similar share of paths, but its single
traveling wave covers less of the real fish's motion (0.74 against 0.85; Figure~\ref{fig:coverage}), so it finishes the task without showing the full range of motion of the real animal.

\subsection{Task 2: Target Reaching}
\label{sec:task2}

\begin{figure}[H]
    \centering
    \includegraphics[width=0.60\linewidth,trim=5bp 89bp 5bp 120bp,clip]{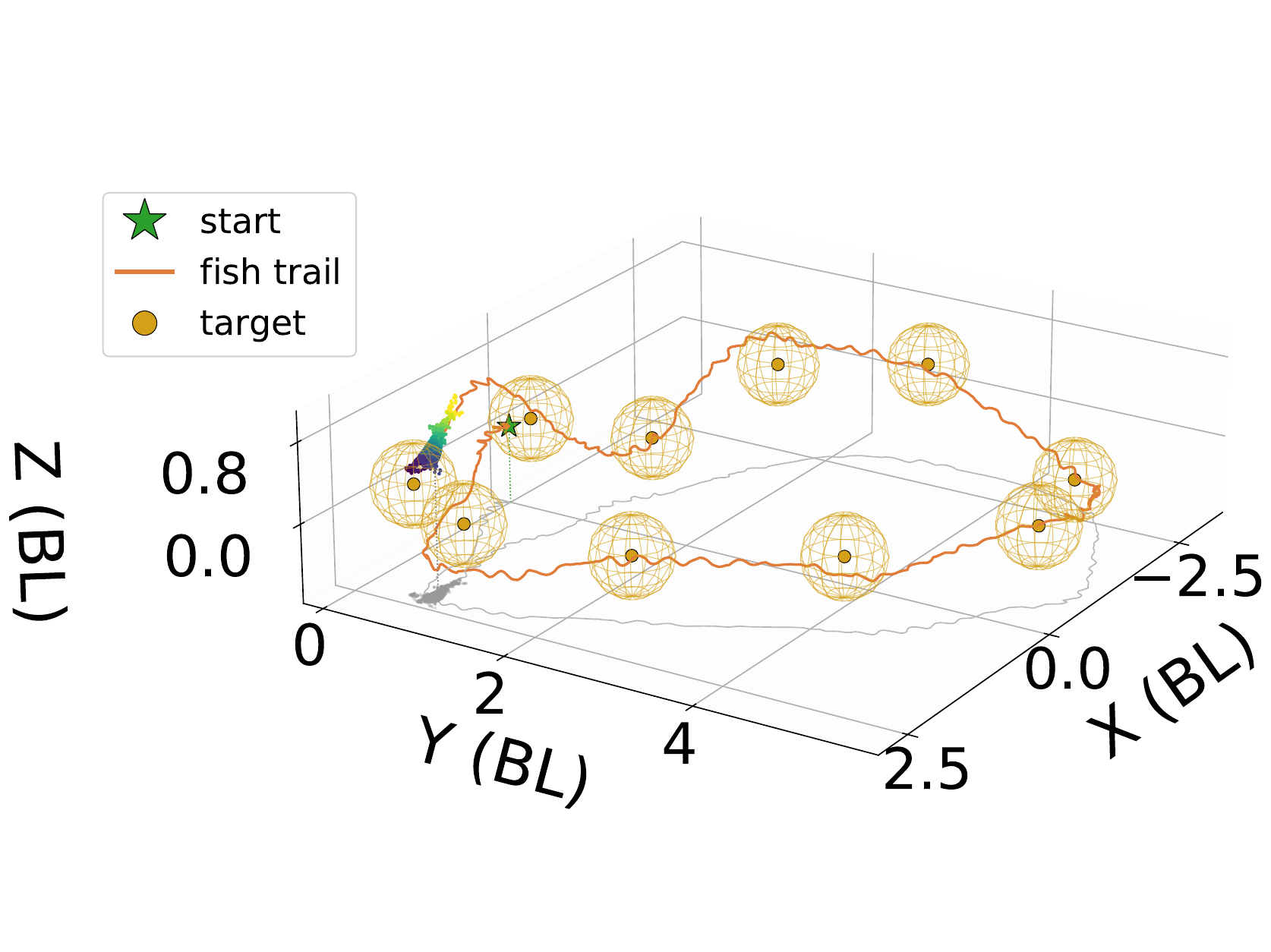}
\caption{One Task~2 episode of BLM+RL (lake sturgeon 016): the fish reaches all ten targets. Orange line: the fish's trajectory; spheres: the reach range of each target (radius 0.4\,BL); star: start. Full videos of all methods: \url{https://reach-10-task.vercel.app/}.}
    \label{fig:reach-episode}
\end{figure}

A useful simulated fish also needs to be versatile, and Task~1 shows only part of this.
Task~2 shows the rest of what the fish can do: swimming straight, turning, climbing, diving
and swimming fast.
In each episode, the fish must reach ten 3D targets in sequence, placed around its current
heading (Figure~\ref{fig:reach-targets}); evaluation uses targets never seen in training. We
report the targets reached, the time to reach and the timeout rate, together with the fidelity
measures of Task~1 (Figure~\ref{fig:reach-episode}; Appendix~\ref{app:task2}).

Table~\ref{tab:task2} shows that BLM+RL reaches the most targets (9.0 of 10) with the lowest
timeout rate, and covers 0.96 of the real fish's bending motions against at most 0.88 for any
baseline. CPG+RL is slightly worse than BLM+RL on the task (8.4 targets) with a slightly lower time to reach, but covers only 0.72
of the real motions, while Joint RL and Joint RL+AMP reach about 3 targets or fewer. Similar to Task~1,
the two low-dimensional action spaces are the most capable, and BLM+RL moves most like the
real fish. Its envelope correlation is lower than in Task~1, and the differences between methods on this measure are within one standard deviation.

\begin{table}[H]
\caption{Task~2 target reaching on twelve fish (two per species): targets reached per episode
(out of 10), time to reach (s), fraction of attempts that time out, motion coverage
and bending-envelope correlation. Mean $\pm$ standard deviation over fish; each fish is first averaged over three seeds.}
\label{tab:task2}
\centering
\small
\begin{tabular}{@{}lccccc@{}}
\toprule
Method & \shortstack{Targets\\reached\\$\uparrow$} & \shortstack{Time to\\reach\\$\downarrow$} & \shortstack{Timeout\\rate\\$\downarrow$} & \shortstack{Motion\\coverage\\$\uparrow$} & \shortstack{Envelope\\correlation\\$\uparrow$} \\
\midrule
Joint RL        & 3.1$\pm$2.6 & 8.0$\pm$2.8 & 0.69$\pm$0.26 & 0.69$\pm$0.17 & 0.36$\pm$0.19 \\
Joint RL + AMP  & 2.3$\pm$2.1 & 8.5$\pm$2.7 & 0.76$\pm$0.21 & 0.63$\pm$0.19 & \textbf{0.48}$\pm$0.18 \\
CPG + RL        & 8.4$\pm$2.5 & \textbf{4.3}$\pm$1.3 & 0.16$\pm$0.25 & 0.72$\pm$0.14 & 0.45$\pm$0.18 \\
BCO + RL        & 6.6$\pm$3.4 & 5.6$\pm$2.1 & 0.34$\pm$0.34 & 0.88$\pm$0.06 & 0.23$\pm$0.21 \\
BLM + RL (Ours) & \textbf{9.0}$\pm$1.5 & 4.6$\pm$1.4 & \textbf{0.10}$\pm$0.15 & \textbf{0.96}$\pm$0.03 & 0.33$\pm$0.16 \\
\bottomrule
\end{tabular}
\end{table}

\subsection{Case Study: BlueROV Capture}
\label{sec:task3}

\begin{figure}[H]
    \centering
    \includegraphics[width=1.0\linewidth]{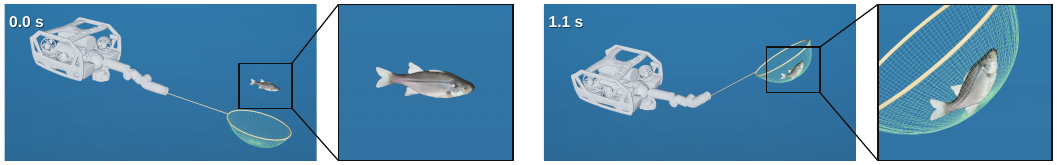}
\caption{Case Study: the ROV scoops up a white bass (white bass 008) with its hand net. Left: start of the episode; right: the fish is inside the net 1.1\,s later. Each close-up shows the fish. Full videos: \url{https://rov-capture.vercel.app/}.}
    \label{fig:rov}
\end{figure}

To further show the practical potential of our dataset and benchmark, we run a case study in which a BlueROV2 with a 6-DoF manipulator and a hand net
learns with PPO to scoop up a freely swimming digital fish, which is driven by its own BLM+RL policy
along paths mimicking its real counterparts from our dataset. We train one capture policy per species
(Table~\ref{tab:task3}; Figure~\ref{fig:rov}; Appendix~\ref{app:rov}).

All six capture policies succeed in at least 98\% of the evaluation episodes (381 of 384 in
total) and converge within 145--217 training epochs. We apply the same task setting and reward function to all six species without per-species tuning, which shows their practical value.

\begin{table}[H]
\caption{Case Study: ROV capture of fish driven by their own Task~1 BLM+RL policy
(64 evaluation episodes per fish).}
\label{tab:task3}
\centering
\small
\begin{tabular}{@{}lcc@{}}
\toprule
Prey & \shortstack{Capture\\success rate\,$\uparrow$} & \shortstack{Epochs to\\convergence} \\
\midrule
Bluegill 020        & 0.98 & 150 \\
Brook trout 019     & 0.98 & 190 \\
Brown trout 015     & 1.00 & 196 \\
Channel catfish 002 & 1.00 & 217 \\
Lake sturgeon 001   & 1.00 & 145 \\
White bass 008      & 0.98 & 152 \\
\bottomrule
\end{tabular}
\end{table}

\subsection{Human Study: Perceived Biological Fidelity}
\label{sec:human-study}

We evaluate the trained digital fish from the practitioners' perspective through a human
study with participants including fish experts at fish hatcheries: twelve raters
score blinded clips of the five methods on held-out Task~1 trajectories. Each clip is rated from 1 to 5, first for how \emph{plausible} the simulated
fish looks on its own, and then, next to the real recording, for \emph{path} match,
\emph{speed} match, \emph{appearance} and overall \emph{similarity}
(Table~\ref{tab:human}; Appendix~\ref{app:human-study}).

The twelve raters judge BLM+RL the most plausible on its own (4.5, against 3.2 for CPG+RL and at
most 2.7 for the joint-space methods). Next to the real recording, BLM+RL
also receives the highest path match and overall similarity (4.1 against 4.0 for CPG+RL),
while CPG+RL is ahead on speed match, so the two low-dimensional methods lead as in
Table~\ref{tab:task1}. The appearance
score does not depend on the method, since all methods use the same mesh; it is 4.9 on
average over all clips, which supports the generated meshes.

\begin{table}[H]
\caption{Human study: expert ratings (1--5, mean $\pm$ standard deviation over the ratings
of twelve raters) of 25 Task~1 clips (3 fish, 5 held-out trajectories); anchors are
zero-action drifting fish. Appearance is the same asset for every method and is reported
once in the text. The videos and questions shown to the raters: \url{https://human-study-rosy.vercel.app/} and Table~\ref{tab:human-items}.}
\label{tab:human}
\centering
\small
\begin{tabular}{@{}lcccc@{}}
\toprule
Method & Plausible\,$\uparrow$ & \shortstack{Path\\match\,$\uparrow$} & \shortstack{Speed\\match\,$\uparrow$} & \shortstack{Overall\\similarity\,$\uparrow$} \\
\midrule
Joint RL        & 2.35 $\pm$ 1.21 & 3.12 $\pm$ 1.05 & 2.38 $\pm$ 1.6 & 3.00 $\pm$ 0.71 \\
Joint RL + AMP  & 2.07 $\pm$ 1.12 & 3.03 $\pm$ 1.23 & 2.45 $\pm$ 1.7 & 2.98 $\pm$ 0.72 \\
CPG + RL        & 3.20 $\pm$ 1.21 & 3.37 $\pm$ 0.91 & \textbf{4.92} $\pm$ 0.38 & 3.95 $\pm$ 0.53 \\
BCO + RL        & 2.73 $\pm$ 1.63 & 2.90 $\pm$ 1.26 & 2.57 $\pm$ 1.62 & 3.02 $\pm$ 0.81 \\
BLM + RL (Ours) & \textbf{4.48} $\pm$ 0.83 & \textbf{3.65} $\pm$ 0.71 & 4.63 $\pm$ 0.75 & \textbf{4.07} $\pm$ 0.51 \\
\midrule
Anchor (drift)  & 2.79 $\pm$ 1.12 & 1.00 $\pm$ 0.00 & 1.00 $\pm$ 0.00 & 1.83 $\pm$ 0.47 \\
\bottomrule
\end{tabular}
\end{table}

\section{Limitations and Discussions}

\textbf{Discussion.} We present Video2SwimFish, a pipeline that turns a video of
one fish into a controllable simulated fish that preserves the morphology and
swimming behavior of the real animal. The released dataset can be used to train
embodied agents to interact with realistic fish, as our Case Study with a BlueROV
illustrates, offering a safer and more scalable alternative to trials with live
fish for aquaculture monitoring and management. The benchmark measures two goals: task
success (does the policy complete the task?) and biological fidelity (does it move like the
fish it was built from?). Our results show that reaching the first goal does not guarantee the
second: CPG+RL completes a similar share of the tasks as BLM+RL, but reproduces less of
the real fish's bending motion. BLM+RL is the best on both goals, but its advantages over a
well-tuned CPG are small compared with the variation across individual fish.
\textbf{Learning a policy that is both task-capable and faithful to the individual animal
therefore remains an open challenge}, and our dataset and metrics can measure progress on it.

\textbf{Limitations.} First, our simulations use an approximate
hydrodynamic model (Appendix~\ref{app:hydro}) rather than full fluid--structure
interaction. Second, we have not tested sim-to-real transfer, owing to limited
access to a test site, robot waterproofing and heat dissipation issues, and
animal-welfare concerns. Third, due to limited compute, we test only 12 of the 120
fish (six in the Case Study).

\section*{Acknowledgments}

This material is based upon work supported by the National Science Foundation
under Grant No.\ 2553466. Any
opinions, findings, and conclusions or recommendations expressed in this
material are those of the authors and do not necessarily reflect the views of
the National Science Foundation.

We thank the North Carolina Wildlife Resources Commission (NCWRC), Inland
Fisheries Division, for supporting this work. We are grateful to David Deaton
and Peter Lamb for hosting our visit to the Marion State Fish Hatchery and for
coordinating our data collection. We also thank Jimmy Lowman, Joshua Sanders,
and the staff of the Table Rock State Fish Hatchery in Morganton; Jeff Evans,
Joshua Cooper, and the staff of the Watha State Fish Hatchery in Pender County;
and Charles Melton and the staff of the Armstrong State Fish Hatchery in
McDowell County for hosting our visits, assisting with data collection, and participating in our human study. We
also thank the Institutional Animal Care and Use Committee (IACUC) of North
Carolina A\&T State University for reviewing and approving our fish video
recording protocol.

\section*{LLM Usage Statement}
We used Qwen3-VL as part of the Video2SwimFish pipeline to select canonical frames and to propose and evaluate fish articulations. We used Meshy to generate 3D fish meshes from the selected frames. These tools contributed to the synthetic fish assets described in Section 3. The authors inspected each generated asset before release. We used Anthropic Claude to assist with editing and refining author-written code.

\bibliography{iclr2027_conference}
\bibliographystyle{iclr2027_conference}
\clearpage
\appendix
\section{Dataset Details}
\label{app:dataset}

\begin{table}[H]
\centering\small
\caption{Dataset summary. Coverage is the fraction of body length spanned by the bone chain; alignment is the mean angle between each bone axis and the local body centerline.}
\label{tab:app-dataset}
\begin{tabular}{lccccccc}
\toprule
Species & $n$ & length [cm] & bones (med.) & coverage & alignment & 3D BLM basis \\
\midrule
Channel catfish & 20 & 12.6--31.1 (med.\ 20.4) & 14.5 & 74\% & $10.9^\circ$ & 20/20 \\
Lake sturgeon  & 20 & 4.6--18.9 (med.\ 10.0)  & 15   & 72\% & $8.7^\circ$  & 18/20 \\
Bluegill       & 20 & 1.6--9.5 (med.\ 6.8)    & 15   & 74\% & $11.1^\circ$ & 18/20 \\
White bass     & 20 & 4.3--12.9 (med.\ 6.3)   & 15   & 74\% & $10.6^\circ$ & 17/20 \\
Brook Trout       & 20 & 5.3--16.7 (med.\ 10.2)    & 14   & 73\% & $10.0^\circ$ & 18/20 \\
Brown Trout     & 20 & 4.7--12.1 (med.\ 9.3)   & 15   & 75\% & $11.3^\circ$ & 17/20 \\
\bottomrule
\end{tabular}
\end{table}

\paragraph{Source videos.}
Each fish is recorded alone in a transparent glass tank (76.2\,cm $\times$ 29.9\,cm $\times$
44.5\,cm, water depth 42\,cm) with two FLIR Blackfly~S BFS-U3-200S6C-C cameras with 12\,mm C-mount
lenses, hardware-synchronized by an external trigger: one top-down, 0.82\,m above the water
surface, and one front-facing, 0.78\,m from the front wall. Camera intrinsics and the
inter-camera geometry are calibrated with a ChArUco board before each session. Recordings are
$2736 \times 1824$ at 32\,fps, two and a half minutes per trial; all behavior signals are resampled to 30\,Hz, the control rate of the simulated fish.
Every video contains exactly one fish; a second silhouette that occasionally appears is the
reflection in the tank wall and is removed by the mesh critic (Appendix~\ref{app:mesh}). The dataset
contains 6 species $\times$ 20 individuals $=120$ fish (Table~\ref{tab:app-dataset}).

\paragraph{Metric scale.}
Pixel-to-meter scale is calibrated once per species from the known 30-inch front
face of the tank ($88.5$, $89.8$, $87.8$, $89.4$, $88.7$ and $89.0$\,px/in for
channel catfish, lake sturgeon, bluegill, white bass, brook trout and brown
trout; $104$\,px/in across the top view). Body length is the median silhouette
length over $\ge 150$ valid frames; frames outside $0.4$--$2.5\times$ the running
median are rejected. When a fish has too few valid frames, the species median
length is used and flagged (\texttt{size\_source}).

\section{Asset Generation Pipeline}
\label{app:pipeline}

All 120 assets were produced by the same automated pipeline
(\texttt{scripts/run\_dataset\_v2sf.py}); no per-fish manual editing is involved. The stages
are run in dependency order and each stage is resumable.

\subsection{Canonical frame and crop}
A VLM (Qwen3-VL-32B-Instruct) selects, from the front-view video, one frame in which the fish
is fully visible and laterally presented. A background-subtraction crop isolates the fish.

\subsection{Image-to-3D and mesh critic}
\label{app:mesh}
The crop is sent to the Meshy-6 image-to-3D API. The returned mesh is validated by a VLM mesh
critic (same model) on four rendered views, rejecting meshes that are not a fish, are
truncated, contain more than one fish, or show the tank rather than the animal. A
deterministic component cleaner then welds coincident vertices, splits connected components,
and removes any secondary component whose oriented bounding box is disjoint from, larger than
10\% outside of, or smaller than 15\% of the area of the main body (this removes the reflection
fish that the image model sometimes hallucinates as a second body). A mesh that fails the
critic is resubmitted once with a corrected crop.

\subsection{Canonicalization}
The mesh is leveled by aligning its first principal axis to $+x$ (head at $+x$), scaled to
the measured body length, and re-centered so the bounding-box center sits at the origin.

\subsection{Skeleton construction: VLM actor--critic with geometric regularization}
\label{app:skeleton}
Articulation is produced by an actor--critic loop over rigid box bones:
\begin{itemize}\setlength{\itemsep}{1pt}
\item \textbf{Actor.} Qwen3-VL-32B-Instruct fine-tuned with LoRA (rank 8, $\alpha$ as in
  \texttt{train\_actor\_lora.py}, 4 epochs, learning rate $10^{-4}$) on 14 expert-annotated
  fish skeletons. Given three orthographic renders of the current skeleton inside the
  semi-transparent skin, it emits one edit per step (\texttt{add}, \texttt{remove},
  \texttt{reposition}, \texttt{resize}) with a position and size in meters.
\item \textbf{Geometric verifier.} Every proposed bone is shrunk anisotropically (per axis)
  until all its vertices lie inside the skin, and snapped to the local body centerline.
\item \textbf{Critic.} The base VLM scores the skeleton from 1 (very poor) to 5 (excellent)
  from the same three views plus measured midline facts (offset of each bone from the
  centerline, coverage, gaps); a skeleton is accepted at score $\ge 4$. The loop runs for at
  most 6 iterations. Deterministic decoding (temperature 0) can fall into add/remove
  2-cycles; after 4 alternating iterations the actor temperature is raised to 0.7 to escape.
\item \textbf{Structural completion $\mathcal{R}$.} The accepted (or best-scoring) skeleton is passed
  through the deterministic structural completion step of Sec.~3.2: bones are
  snapped to a thickness-weighted body centerline, overlapping bones are merged, inner gaps
  wider than 2\% of body length are filled with box bones of the same section, and the chain
  is extended toward the skull and the caudal-fin base (the points where the cross-section
  falls below 15\% / 3\% of its maximum), with a cap of 22 bones. Every new or moved bone
  goes through the same containment shrink, and the critic re-scores the result.
\end{itemize}
The actor typically places 5--6 bones (median 6), as in the expert skeletons it was fine-tuned
on (4--8 bones, median 6). The structural completion step then adds 6--14 bones (median 10)
by filling gaps and extending the chain toward the skull and the caudal-fin base, so the
exported chains have 12--19 bones (median 15), 1755 bones in total. The actor thus decides
where the articulated region lies and which bone shape it uses; the completion step makes the
chain dense and long enough for simulation. The final critic score is 5 for 114/120 fish.

\subsection{Export and Stage-1 verification}
Bones become rigid cylinders/boxes connected by 3-DoF D6 joints ($\pm 80^\circ$ per axis,
PD position drives, stiffness $k_p = 240$, damping $6$); the skin becomes a PhysX FEM
deformable body attached to the bones. Each asset is checked for (i) containment: no more
than 5 bone vertices outside the skin and penetration $<2$\,mm; (ii) coverage; and
(iii) alignment: mean bone--centerline angle $<15^\circ$, max $<30^\circ$. 87/120 assets pass
the strict containment test; the remainder have 1--5 vertices marginally outside (one has 27).
Coverage is bounded by design: the expert skeletons used for training span only 38--64\% of
body length (no bones in the skull or the caudal fin), and the regularized chains reach
64--83\%.

\subsection{Hydrodynamics}
A panel model replaces the CFD: the FEM skin is partitioned into per-bone slices; each slice
receives quadratic normal drag ($C_{d,\perp}=1.0$, the source of tail thrust) and skin
friction ($C_{d,\parallel}=0.01$), applied as an external wrench every physics step, together
with the reactive (added-mass) force of elongated-body theory. Per-bone force and torque
clips guard against numerical runaway (details in Appendix~\ref{app:hydro}).

\subsection{Behavior extraction}
\label{app:behavior}
The top-view video is segmented by background subtraction inside the tank interior (which
excludes cables and walls outside the water), with a low contrast floor so that fish over dark
water are kept. The midline is the skeleton of the darkest 75\% of the fish's pixels, which
removes shadows attached to the body, extended along its tangent to the ends of the silhouette;
it is resampled to 20 arc-length stations, and the signed curvature $\kappa\cdot\mathrm{BL}$ is
computed per frame. Head and tail are assigned by a small CNN that recognizes the snout and the
tail from image patches, smoothed over time by a Viterbi pass. A per-frame quality gate (a
logistic regression on features of the automatic result only, such as maximum curvature,
minimum width and length deviation) rejects likely failures, and a video in which the snout
cannot be recognized in most frames is marked unusable. The front-view video is processed
with the same steps, plus three that use the top view: the top-view position selects which
blob is the fish (so a second fish in the tank is not tracked), the
top-view length bounds the front-view length, and frames in which the fish faces the camera
are discarded, since the side view then shows no body shape. The two midlines are fused into a 3D midline (longitudinal and lateral coordinates from the top view, vertical from the front view), head-anchored and normalized by body length; frames in which either view fails or the head/tail identity disagrees between views are discarded. The fish's bone chain is fitted to each 3D midline by exact sequential inverse kinematics (yaw then pitch per joint, roll fixed to zero), and a per-fish PCA of the block-normalized joint angles $[\psi\,|\,\theta]$ gives the BLM basis (Sec.~3.3). The top-view curvature $\kappa\cdot\mathrm{BL}$ is kept only for the fidelity metrics (motion coverage and envelope correlation). The same frames provide the 3D reference trajectories (horizontal position and heading from
the top view, depth from the front view; 3\,s windows, re-rooted at the window start) used by
the trajectory-following task, and the curvature used by the biofidelity metrics. Windows from
a few whole recording runs of each fish are held out as the evaluation split
(Appendix~\ref{app:blm-split}).

\subsection{Data split and what each motion prior sees}
\label{app:blm-split}
Each recording is cut into runs (continuous segments of valid frames). A few whole runs of
each fish are held out, and all evaluation windows of Task~1 come from them. No method sees
the held-out trajectories during RL training. The motion priors, however, are fitted on
different frames. The AMP reference and the BCO demonstrations use the training runs only.
The BLM (PCA mean, basis and the two $P_{99}$ bounds) and the CPG parameters (amplitude
envelope, frequency range and phase) are fitted once per fish on all valid frames of the
recording, which include the held-out runs. Both are unsupervised summaries of body shape and
contain no path, speed or reward. Still, the held-out body shapes did enter the BLM, so we
measure how much this matters. Without retraining, we refit the BLM of each of the twelve
benchmark fish on its training runs only and compare it with the released one, using the
eight modes that the policies act on (Table~\ref{tab:blm-trainonly}).

\begin{table}[H]
\centering
\footnotesize
\setlength{\tabcolsep}{4pt}
\caption{BLM fitted on the training runs only against the released BLM (all frames), eight
modes. Angle: principal angles between the two subspaces in joint-angle space (mean over the
eight modes; largest of the four leading modes). Variance: share of the joint-angle variance
of all frames inside each subspace. Bounds: change of the $P_{99}$ bounds when only training
frames are used (median over modes). Inside: share of held-out frames whose coefficients lie
inside the bounds. Error: mean reconstruction error of the held-out midlines.}
\label{tab:blm-trainonly}
\resizebox{\linewidth}{!}{%
\begin{tabular}{lcccccccc}
\toprule
Fish & \shortstack{Held-out\\frames (\%)} & \shortstack{Angle,\\mean ($^\circ$)} & \shortstack{Angle, 4\\modes ($^\circ$)} & \shortstack{Variance\\all / train} & \shortstack{$a_{\max}$\\change (\%)} & \shortstack{$\Delta a_{\max}$\\change (\%)} & \shortstack{Inside\\all / train} & \shortstack{Error ($10^{-3}$\,BL)\\all / train} \\
\midrule
Bluegill 015 & 38 & 1.9 & 2.7 & 0.996 / 0.996 & 8.3 & 5.9 & 0.99 / 0.99 & 0.4 / 0.5 \\
Bluegill 020 & 22 & 1.5 & 5.1 & 0.996 / 0.996 & 2.5 & 5.4 & 0.97 / 0.97 & 0.5 / 0.7 \\
Brook trout 019 & 39 & 5.4 & 3.3 & 0.997 / 0.997 & 2.8 & 3.0 & 0.95 / 0.95 & 0.8 / 1.2 \\
Brook trout 020 & 27 & 9.4 & 2.2 & 0.994 / 0.996 & 5.1 & 3.4 & 0.94 / 0.94 & 0.9 / 0.9 \\
Brown trout 011 & 33 & 1.7 & 3.1 & 0.998 / 0.998 & 5.7 & 7.8 & 0.98 / 0.99 & 0.3 / 0.4 \\
Brown trout 015 & 40 & 3.8 & 3.2 & 0.999 / 0.998 & 4.9 & 9.7 & 0.96 / 0.94 & 0.2 / 0.3 \\
Channel catfish 001 & 19 & 1.6 & 1.1 & 0.995 / 0.995 & 4.4 & 4.7 & 0.93 / 0.93 & 1.4 / 1.6 \\
Channel catfish 002 & 5 & 0.8 & 0.8 & 0.995 / 0.995 & 1.6 & 0.4 & 0.92 / 0.92 & 0.9 / 1.0 \\
Lake sturgeon 001 & 9 & 0.6 & 0.8 & 0.996 / 0.996 & 1.7 & 1.3 & 0.97 / 0.97 & 1.0 / 1.1 \\
Lake sturgeon 016 & 16 & 1.6 & 1.9 & 0.993 / 0.993 & 2.1 & 5.7 & 0.92 / 0.89 & 1.7 / 1.8 \\
White bass 008 & 18 & 1.4 & 3.5 & 0.998 / 0.998 & 9.9 & 6.0 & 0.85 / 0.79 & 0.5 / 0.5 \\
White bass 020 & 12 & 0.8 & 3.5 & 0.997 / 0.996 & 2.8 & 1.8 & 0.98 / 0.97 & 0.7 / 0.8 \\
\midrule
Mean & 23 & 2.5 & 2.6 & 0.996 / 0.996 & 4.3 & 4.6 & 0.95 / 0.94 & 0.8 / 0.9 \\
\bottomrule
\end{tabular}}
\end{table}

\emph{Subspace.} The two subspaces differ by a mean principal angle of $2.5^\circ$, and by at
most $5.1^\circ$ for the four leading modes. The last, low-variance modes rotate more in three
fish (largest angle $18^\circ$ to $71^\circ$), but both subspaces contain the same share of
the joint-angle variance of all frames (0.996).
\emph{Bounds.} In the released basis, the bounds computed from the training frames alone
differ from the released ones by 4.3\% ($a_{\max}$) and 4.6\% ($\Delta a_{\max}$) on average;
94\% of the held-out frames lie inside the train-only bounds, against 95\% for the released
bounds.
\emph{Held-out shapes.} The train-only basis reconstructs the held-out midlines with a mean
error of $0.9\times10^{-3}$\,BL, against $0.8\times10^{-3}$\,BL for the released basis.
A BLM built without the held-out runs is therefore almost the same action space. We did not
retrain the policies with the train-only BLM, so the motion coverage and envelope correlation
of BLM+RL and CPG+RL may still be slightly optimistic; completion rate, Fr\'echet distance and
speed error are computed from paths, which neither prior contains.

\subsection{Midline extraction accuracy}
\label{app:midline-acc}
We compare the automatically extracted midlines with the human annotations of
Section~\ref{sec:data-collection}. Annotators click the snout, 3--6 points along the midline,
the caudal peduncle and the tail tip in synchronized top and front frames, without seeing the
automatic result; frames are sampled per fish uniformly in time, at high body curvature, and
in short bursts. We score every annotated frame that the full extraction pipeline above keeps,
including its quality gate, in each view. \emph{Chamfer distance} is the mean symmetric
distance between the automatic and human midlines, and \emph{length error} is the absolute
difference of their snout-to-peduncle lengths; both are in percent of the fish's standard
length. The learned parts of the pipeline (the head/tail CNN and the quality gate) are
evaluated leave-one-fish-out, so no fish is scored by a model trained on it. Each fish
contributes the median over its frames, and we report the median over fish with the range
across fish. As a noise floor, a subset of frames was annotated twice.

\begin{table}[H]
\centering
\small
\setlength{\tabcolsep}{2pt}
\caption{Midline extraction accuracy against human annotations (\% of standard length; median
over fish, range in brackets). Human: Chamfer\,/\,length error between repeated annotations of
the same frames.}
\label{tab:midline-acc}
\begin{tabular}{lcccc}
\toprule
View & \shortstack{Fish /\\frames} & \shortstack{Chamfer\\distance} & \shortstack{Length\\error} & \shortstack{Human\\Chamfer /\\length} \\
\midrule
Top & 11 / 286 & 0.9 [0.5, 1.2] & 9.3 [3.5, 13.0] & 0.7 / 1.5 \\
Front & 12 / 196 & 5.0 [1.6, 11.0] & 16.4 [7.3, 24.0] & 1.6 / 4.1 \\
\bottomrule
\end{tabular}
\end{table}

In the top view, the extracted midline follows the human one closely on every fish (Chamfer
0.5--1.2\%), close to the human noise floor of 0.7\%. One of the twelve annotated fish, a brown
trout recorded over dark water where it is barely visible, is marked unusable by the video
check and is not scored in the top view. The front view is less accurate (Chamfer 5.0\%), since
the side silhouette includes the fins and reflections from the tank floor and side glass, and
the peduncle is hard to locate even for annotators (human floor 1.6\%). Length errors are larger
than shape errors in both views (9.3\% top, 16.4\% front) and come mainly from where the
midline ends at the tail. Because each midline is normalized by its own length before the
joint fit, a length error shifts the positions along the body at which the bone directions are
read rather than adding bending that is not there; it nevertheless blurs where along the body
a bend occurs, most for the vertical bending from the front view.

\section{Mesh Fidelity: Silhouette IoU Against Hand-Drawn Outlines}
\label{app:mesh-iou}

A 3D ground truth of a live fish is not available, so we test the generated meshes
(Sec.~\ref{app:mesh}) in 2D: we compare the silhouette of each mesh with the outline of the
real fish in two orthogonal camera views. The \emph{side view} is the canonical frame that
the mesh was generated from. The \emph{top view} comes from the second camera and was never
given to the image-to-3D model, so it tests the part of the shape that the input image does
not show, i.e.\ the body width and the fins seen from above.

\paragraph{Fish outlines.}
The outline of the real fish is drawn by hand. Two annotators trace the fish in each image
as a polygon in a web tool (body and all visible fins, including translucent ones); no
automatic segmentation is involved. The tool hides the fish identity and never shows the
mesh. For the top view we take one frame of the top camera, at least 5\,s away from the
canonical frame, in which the body is straight: the frame with the lowest mean curvature of
the automatic midline over a 0.4\,s window, confirmed by eye. The mesh is rigid, so a bent
body would lower the IoU for a reason unrelated to the mesh.

\paragraph{Mesh silhouettes.}
The mesh silhouette is computed from the geometry alone: every triangle of the mesh is
projected orthographically, along the lateral axis for the side view and along the vertical
axis for the top view, and filled. No image is rendered and no threshold is applied.

\paragraph{Alignment.}
The two silhouettes are brought into a common frame by a similarity transform computed from
image moments (centroid, principal axis, extent along the axis, i.e.\ 4~DOF). The outlines
carry no head label, so we keep the better of the four flips (head/tail and up/down). On top
of this we report a 6-DOF affine refinement (anisotropic scale, rotation, shear, translation;
Powell search maximising IoU), which absorbs the mild perspective of the photograph. All
alignments are global 2D transforms; no non-rigid warping is used.

\begin{figure}[H]
\begin{center}
\includegraphics[width=\linewidth]{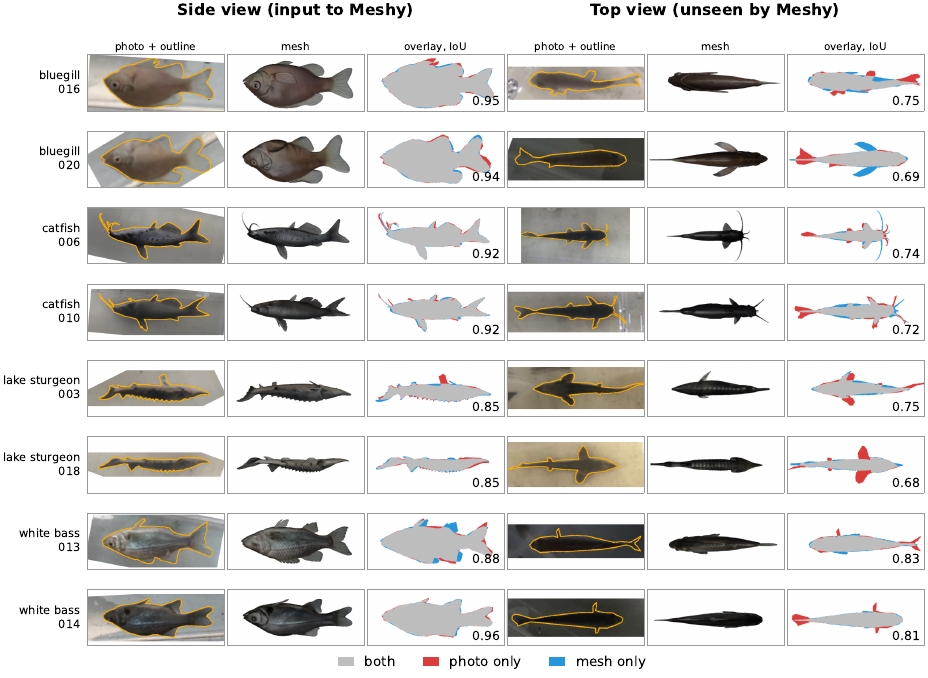}
\end{center}
\caption{Mesh-versus-photo silhouettes for eight of the fish of Table~\ref{tab:canonical-iou-best8},
in the side view that the mesh was generated from (left) and in the top view that the
image-to-3D model never saw (right). Each group shows the photograph with the hand-drawn
outline, the mesh from the same direction, and the two silhouettes after affine alignment
with their IoU (gray: both, red: photo only, blue: mesh only).}
\label{fig:mesh-iou}
\end{figure}

\paragraph{Fish.}
Table~\ref{tab:canonical-iou-best8} reports all 120 fish of the dataset.
Fig.~\ref{fig:mesh-iou} shows eight of them, two from each of four species, chosen for a
straight body in both views and not by their IoU; the IoU of each of these fish is printed
in the figure.

\paragraph{Results.}
Over all 120 fish, the affine IoU is $0.91 \pm 0.19$ in the side view and $0.75 \pm 0.16$ in
the unseen top view (Table~\ref{tab:canonical-iou-best8}). The eight fish of
Fig.~\ref{fig:mesh-iou} have the same means ($0.91$ and $0.75$), so they are representative
of the dataset. Stripping the fins from both silhouettes with a morphological opening
(kernel $6\%$ of the body length) changes the means by at most $0.01$.

The lateral profile, which the model saw, is reproduced closely; the remaining disagreement
comes from the pose of the fins and the barbels. The top view is less accurate. The
overlays in Fig.~\ref{fig:mesh-iou} show three sources: the pectoral fins of the mesh are
spread while the swimming fish often holds them against the body, the tail of the real fish
is rarely perfectly straight, and the body width of the mesh differs from that of the animal.
The affine fit confirms the last point: for the eight fish of Fig.~\ref{fig:mesh-iou}, it
changes the two scales by at most $9\%$ in the side view (rotation $<2^\circ$, shear
$<0.2$), but by up to $21\%$ in the top view. The
image-to-3D model has to guess the body width from a single lateral image, and this is the
least accurate dimension of our meshes.

\begin{table}[H]
\centering
\small
\caption{Silhouette IoU between the generated mesh (orthographic projection of the mesh triangles) and hand-drawn outlines of the real fish: mean $\pm$ standard deviation over all 120 fish of the dataset. Side view: the canonical input frame; top view: a frame of the second camera that the image-to-3D model never saw. Similarity: 4-DOF alignment; affine: 6-DOF refinement; body only: both silhouettes with fins morphologically stripped before the affine IoU.}
\label{tab:canonical-iou-best8}
\begin{tabular}{lccc}
\toprule
View & \shortstack{IoU\\(similarity)} & \shortstack{IoU\\(affine)} & \shortstack{IoU\\(body only)} \\
\midrule
Side view (input) & 0.85$\pm$0.28 & 0.91$\pm$0.19 & 0.92$\pm$0.15 \\
Top view (unseen) & 0.69$\pm$0.18 & 0.75$\pm$0.16 & 0.75$\pm$0.20 \\
\bottomrule
\end{tabular}
\end{table}

\section{Simulation and Control}
\label{app:sim}
Isaac Sim 5.1 / Isaac Lab (PhysX, GPU). Physics step $1/120$\,s, control at 30\,Hz
(decimation 4), zero gravity (neutral buoyancy), FEM material Young's modulus $10^5$\,Pa,
Poisson ratio $0.45$, elasticity damping $0.05$. 256 parallel environments per run. Every fish uses the
same actuator preset (3-DoF joints, $\pm80^\circ$, PD stiffness 240) without any per-fish
adjustment, and the same fluid-force model (Appendix~\ref{app:hydro}) for every method.

\paragraph{Action spaces.}
\emph{Joint}: PD position targets for the lateral and vertical DoF of every joint, with a
bounded rate (at most $5^\circ$ per step and $|q|\le10^\circ$ per joint; the summed bend of each
family at most $90^\circ$, changing by at most $40^\circ$ per step).
 \emph{BLM}: coefficients of the per-fish joint-angle PCA basis (8 modes), decoded directly to joint targets $\mathbf{q}^{*}=\boldsymbol{\mu}+\Sigma V\mathbf{a}$; the yaw block drives the lateral DoF and the pitch block the vertical DoF of each joint. The only per-fish calibration is the sign of the simulator joint axes relative to the IK angles, obtained by correlating the IK yaw with the lateral bending the simulator produces for the same frames; the pitch block uses the same sign, since a vertical mirror of the body shape is hydrodynamically equivalent in neutrally buoyant water. With $\hat{\mathbf{a}}_t\in[-1,1]^d$ the normalized policy action, the coefficients are updated incrementally,
\begin{equation}
    \mathbf{a}_{t+1}
    =
    \operatorname{clip}\left(
        \mathbf{a}_t + \hat{\mathbf{a}}_t\odot\Delta\mathbf{a}_{\max},\;
        -\mathbf{a}_{\max},\;
        \mathbf{a}_{\max}
    \right),
\end{equation}
which keeps the rate of change of every coefficient inside the range observed in the fish's own video; the joints then track $\mathbf{q}^{*}_{t+1}$ with PD control, so no curvature-to-joint controller has to be identified.
\emph{CPG}: a traveling-wave generator with amplitude envelope, frequency and phase offsets
fitted to the fish's own curvature (parameters in \texttt{cpg\_params.npz}); the policy
outputs $[\Delta A, \Delta f, \Delta b]$ for the lateral wave plus a static vertical bend.

\subsection{BLM-to-joint mapping accuracy}
\label{app:blm-tracking}
We test whether the BLM action space is realized by the simulated fish, independently of any
policy. For each of the twelve benchmark fish, the real fish's own BLM coefficients
(continuous recordings of at least 2\,s, gaps of up to 3 frames interpolated, resampled from
16 to 30\,Hz) are fed open-loop as the action of the Task~1 environment, with the same PD
gains, physics step, hydrodynamics, coefficient limits and joint limits as in training;
64 environments run for 330 control steps, and the first second after every reset is
discarded. The error from the real IK joint angles to the joint angles the simulator realizes
is split into three stages: (1) truncation to 8 modes, (2) the coefficient and joint limits,
and (3) PD tracking. We measure it on joint angles, on the curvature $\kappa\cdot\mathrm{BL}$
at each joint, and on the 3D midline obtained by forward kinematics of the bone chain
(mean point distance in BL).

\begin{table}[H]
\centering
\small
\setlength{\tabcolsep}{3pt}
\caption{BLM-to-joint mapping accuracy on the twelve benchmark fish (mean $\pm$ std over fish).
Joint $R^2$, RMSE and amplitude ratio compare the realized with the commanded joint angles;
curvature $R^2$ compares the input and output of each stage alone, and the real IK curvature
with the realized one end to end.}
\label{tab:blm-tracking}
\begin{tabular}{lcc}
\toprule
 & Lateral & Vertical \\
\midrule
Joint $R^2$ & $0.992\pm0.005$ & $0.993\pm0.001$ \\
Joint RMSE ($^\circ$) & $0.25\pm0.09$ & $0.23\pm0.10$ \\
Command std ($^\circ$) & $2.95\pm0.79$ & $2.65\pm1.12$ \\
Amplitude ratio & $0.982\pm0.011$ & $0.985\pm0.003$ \\
Lag (ms) & 0 & 0 \\
\midrule
Curvature $R^2$: truncation & $0.997\pm0.002$ & $0.998\pm0.001$ \\
\quad limits & $0.999\pm0.001$ & $0.999\pm0.001$ \\
\quad PD tracking & $0.992\pm0.005$ & $0.993\pm0.001$ \\
\quad end to end & $0.988\pm0.005$ & $0.990\pm0.002$ \\
\bottomrule
\end{tabular}
\end{table}

The joints follow the command closely: the RMSE is about a tenth of the command's own
standard deviation, the amplitude is preserved to within 2\%, and there is no lag at the
33\,ms control resolution. The coefficient limits are active on only $2.2\pm1.1$\% of the
steps. In midline terms, truncation moves the 3D midline by $0.001$\,BL and the whole chain
(truncation, limits and tracking) by $0.007\pm0.002$\,BL, or $0.016\pm0.005$\,BL at the tail
tip. The lowest fish (lake sturgeon 001) still reaches a lateral joint $R^2$ of $0.981$ and an
end-to-end curvature $R^2$ of $0.976$. The BLM therefore loses almost nothing between the real
body shape and the simulated one; differences in the learned gaits come from the policy, not
from the mapping.

\section{Benchmark Tasks}
\label{app:tasks}

\subsection{Task 1: trajectory following}
\label{app:task1}
Each episode samples a 3\,s window of the fish's own real 3D trajectory (position and heading),
re-rooted at the fish's reset pose (translation plus a rotation about the vertical axis); the
fish starts at rest. The phase $s_t$ is the arc length of the nearest path point to the fish;
the look-ahead target $p^*_t$ sits $0.3$\,BL ahead. The per-step reward is
$0.3\,e^{-2d} + 0.3\cos\Delta\theta + 0.1\,r_v + 1.0\,\mathrm{clip}(v_\parallel,0,2) - 0.3
- 0.01\,r_a$, with $d$ the distance to $p^*_t$ (BL), $\Delta\theta$ the heading error to the
path tangent (to $p^*_t$ when the fish is more than 1\,BL off the path), $r_v=-|v-v_{\mathrm{ref}}|$
the speed-match term (BL/s), $v_\parallel$ the speed along the path (BL/s, paid only within
1\,BL of the path) and $r_a=0.01\,\|a_t\|^2$ the action energy; completion pays a bonus of 20 and an out-of-bounds
termination a penalty of 180. The episode is completed when the phase passes 90\% of the path
and the fish is within $0.5$\,BL of the path end; it also ends after 20\,s.

\paragraph{Evaluation.} The checkpoint with the best training completion is evaluated on 256
episodes, one per environment, each replaying a held-out window (environment $e$ uses window
$e \bmod K$, the same for every method, seed and fish), with stochastic actions as in training;
an episode that does not finish counts as not completed. Metrics:
\begin{itemize}\setlength{\itemsep}{1pt}
\item \emph{completion}: fraction of episodes that complete the path;
\item \emph{Fr\'echet distance}: discrete Fr\'echet distance between the executed and
  reference 3D paths in body lengths;
\item \emph{motion coverage} (the recall of \citet{kynkaanniemi2019improved}): the curvature $\kappa\cdot\mathrm{BL}$ at
  20 stations of the first episode of every environment is resampled to 16\,fps and cut into
  8-frame (0.5\,s) windows; a real window counts as covered when it lies within the
  $k$-th-nearest-neighbour radius ($k=5$) of some simulated window, and coverage is the covered
  fraction of the real windows;
\item \emph{envelope correlation}: the curvature of the longest clean stretch of each
  episode is high-pass filtered and Hilbert-transformed, and the amplitude of its dominant
  complex mode along the 20 stations is the bending envelope; we report its Pearson
  correlation with the real fish's envelope (scale-free, so a small-amplitude gait cannot
  score by being small);
\item \emph{speed error} $\Delta v$: absolute difference between the simulated and real mean
  swimming speed (BL/s).
\end{itemize}
Each fish is averaged over its three seeds before the mean and standard deviation over fish
are taken.

\paragraph{Baselines.} Joint RL (PPO on joint targets), Joint RL+AMP (the Joint RL task plus an
adversarial motion prior whose features are the lateral and vertical joint angles, the speed
and the nose direction, compared with the same quantities of the fish's real IK joint angles;
discriminator 2$\times$128, learning rate $1.5\times10^{-5}$, update every 128 steps, batch
512, input noise 0.3, replay 40k, style weight 0.2), CPG+RL, BCO+RL (the Joint RL task, with
PPO initialized from a behavior-cloned policy whose pseudo-labels come from an inverse
dynamics model trained on the fish's own random-interaction data) and BLM+RL (PPO in the BLM
space). All controllers share the task, reward and observations; only the action space
differs.

\paragraph{PPO.} rl\_games; actor--critic MLP $[128,64]$ with ELU, state-independent
log-std initialized at $-0.5$, $\gamma=0.99$, GAE $\lambda=0.95$, learning rate
$3\times10^{-4}$ with KL-adaptive schedule (target 0.008), clip 0.2, entropy 0.003, horizon
256, 5 mini-epochs, minibatch 32 per environment, observation and value normalization.
All baselines share these settings and the same fish, actuator gains, joint limits and
reference windows.

\subsection{Task 2: target reaching}
\label{app:task2}
Task~2 tests whether a controller can swim toward goals that do not come from the fish's own
recordings. It reuses the fish, controllers, observations (except the task features), PPO
settings and stopping rule of Task~1; only the target, the reward and the episode structure
change.

\paragraph{Episode structure.} An episode consists of exactly ten \emph{attempts}, one target
each; the attempt index is not observed. An attempt succeeds when the 3D distance between the
fish's root and the target falls below $0.4$\,BL. It fails when 20\,s pass without a reach
(\emph{timeout}) or when the fish is more than 6\,BL from the target (\emph{abort}). Either way
the next target is drawn at once from the fish's current pose; the fish is never reset inside
an episode, so it must recover from its own failures. The episode ends after the tenth attempt
(at most 200\,s, capped at 205\,s). It ends early, and the remaining attempts count as missed,
when the fish leaves a 60\,BL region around its spawn point.

\begin{figure}[H]
\centering
\includegraphics[width=0.85\linewidth]{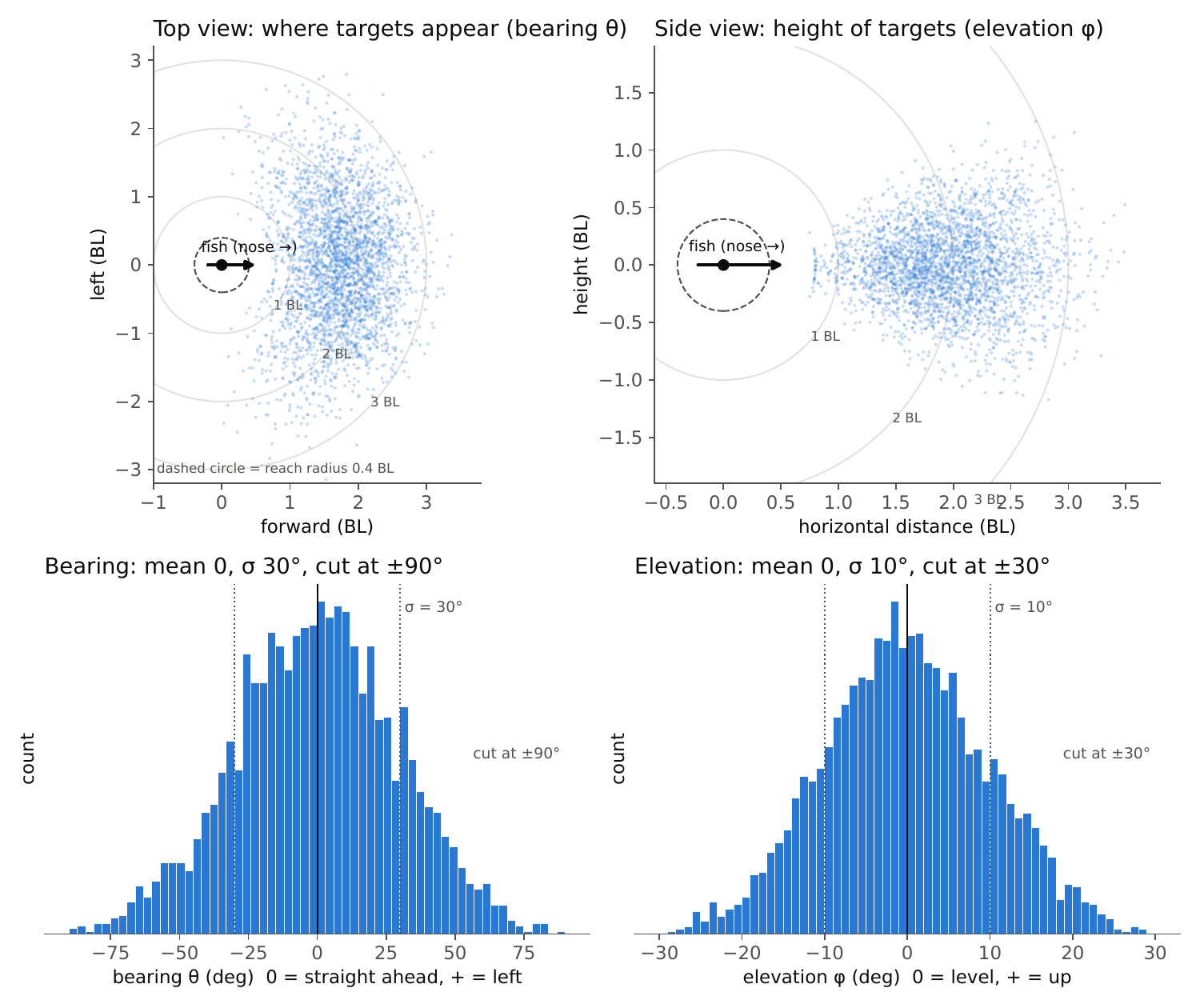}
\caption{Task~2 target distribution, 3000 targets sampled with the evaluation settings. The
fish is at the origin with its nose along $+x$. \emph{Top left:} top view; the dashed circle
is the reach radius (0.4\,BL) and the gray rings mark 1, 2 and 3\,BL. \emph{Top right:} side
view, horizontal distance against height. \emph{Bottom:} histograms of the bearing $\theta$
(mean $0^\circ$, $\sigma=30^\circ$, truncated at $\pm90^\circ$) and the elevation $\phi$
(mean $0^\circ$, $\sigma=10^\circ$, truncated at $\pm30^\circ$). Most targets lie ahead of the
fish at about its depth, with occasional sharp turns and climbs.}
\label{fig:reach-targets}
\end{figure}

\paragraph{Target distribution.} Each target is placed at
$p + d\,[\cos\phi\cos(\alpha+\theta),\ \cos\phi\sin(\alpha+\theta),\ \sin\phi]$, where $p$ is
the fish's position and $\alpha$ its nose azimuth (heading in the horizontal plane) when the
target is drawn. The bearing $\theta$ is measured from the nose heading and the elevation
$\phi$ from the horizontal plane:
$\theta\sim\mathcal{N}(0,30^\circ)$ truncated to $\pm90^\circ$,
$\phi\sim\mathcal{N}(0,10^\circ)$ truncated to $\pm30^\circ$ (both by inverse-CDF sampling),
and $d\sim\mathcal{N}(2,0.5^2)$\,BL clipped to $[0.8,3.6]$\,BL. Both truncations sit at
$3\sigma$, so they only remove extreme draws. Figure~\ref{fig:reach-targets} shows the
result: most targets lie 1.5--2.5\,BL ahead of the nose at about the fish's depth; about 4\%
require a turn of more than $60^\circ$ and about 4\% a climb or dive of more than $20^\circ$.
The minimum distance of 0.8\,BL keeps every new target outside the reach radius. The
distribution is the same for every fish and, since all lengths are in body lengths, the task
has the same difficulty for small and large fish.

\paragraph{Shared target sequence.} The four uniform draws of attempt $k$ in episode $n$ of
environment $e$ (bearing, elevation and two for the distance) are a splitmix64 hash of
$(\text{seed}, e, n, k)$, not the output of a stateful random generator. The targets therefore
do not depend on how the fish swims or on when earlier episodes ended: every method and every
training seed of a fish faces the same sequence of relative targets. Training uses seed 0 and
evaluation uses seed 1000, so no evaluation target is seen in training.

\paragraph{Observation and reward.} The policy observes the unit direction to the target in its
body frame and the distance in BL, in place of the path features of Task~1; all other
observations are those of Task~1. The per-step reward is
$30\,g(\Delta d) + \max(\cos H,0)\,\mathrm{clip}(v_{\to},0,0.5)
+ 25\,\mathbb{1}[\mathrm{reach}]\max(\cos H,0) - 0.001\,\overline{\dot q^{2}}$, where
$\Delta d$ is the per-step progress toward the target in BL, clipped to $[-1,1]$; $g$
multiplies its positive part by the nose alignment $\max(\cos H,0)$, with $H$ the angle
between the nose and the direction to the target, and keeps its negative part unchanged;
$v_{\to}$ is the speed toward the target in BL/s; and the last term is an energy cost on the
joint velocities. The alignment factors pay only for head-first approaches, so a fish cannot
score by drifting or swimming backward into the target. The reward is the same for all five
controllers (Joint RL+AMP adds its style term as in Task~1).

\paragraph{Training and evaluation.} Every method is trained on each of the twelve fish with
three seeds. The success signal for the stopping rule of Task~1 is the per-attempt reach rate
over the most recent 4096 finished attempts, since full episodes are too long to give a
timely signal. The checkpoint with the best reach rate is evaluated on 256 environments, one
episode (ten attempts) each, with stochastic actions as in training and target seed 1000.

\paragraph{Metrics.}
\begin{itemize}\setlength{\itemsep}{1pt}
\item \emph{targets}: targets reached per episode, out of 10 (missed attempts after an early
  termination count as failures);
\item \emph{time to reach}: mean duration of the successful attempts (s);
\item \emph{timeout}: fraction of the played attempts that end in a timeout;
\item \emph{motion coverage} and \emph{envelope correlation}: computed as in Task~1, from the
  body curvature of the first episode of every environment, against the same real recordings
  of the fish.
\end{itemize}
Table~\ref{tab:task2} reports the mean and standard deviation over the twelve fish, each first averaged over its three seeds.

\subsection{Case Study: ROV capture}
\label{app:rov}
A BlueROV2 with a 6-DoF manipulator that holds a hand net (a hoop with a bag) is trained
(PPO, 512 environments, same hyper-parameters) to scoop up a free-swimming fish and lift it.
The prey is driven by the fish's own frozen Task~1 BLM+RL policy (seed 0, stochastic actions).
Whenever it finishes a window, the next real window of the same fish is planted at its current
pose, so that it keeps swimming for the whole 45\,s episode inside a $2.5$\,BL region. Each
episode starts with the net below the fish. Success requires the net to rise by $0.5$\,m with
the fish inside it. Capture policies are trained for six fish, one per species (bluegill 020,
brook trout 019, brown trout 015, channel catfish 002, lake sturgeon 001, white bass 008), with
the same settings for all of them. Training stops when the smoothed success rate stays at or
above $0.95$ for 10 epochs. Table~\ref{tab:task3} reports the epoch at which training stopped
and the success rate of the final policy over 64 evaluation episodes with deterministic
actions (one training run per fish).

\subsection{Training protocol and stopping rules}
\label{app:protocol}
A job scheduler runs several jobs per GPU and stops a run only on convergence: the 10-epoch
smoothed success (completion for Task~1, per-attempt reach rate for Task~2, capture success for
the Case Study) at or above $0.95$ for 10 epochs; or a plateau (no improvement above $0.02$
in the best smoothed success over 100 epochs after at least 150 epochs); at most 800 epochs.
Every method is trained with three seeds (0, 1, 2). The convergence statistics of Section~\ref{sec:task1} are computed over the 36 training runs of each method (twelve fish, three seeds) from the 10-epoch smoothed training completion: the epochs to a completion of 0.5 are counted to the first epoch at which the smoothed value reaches 0.5, and we report the median over the 36 runs (a run that never reaches 0.5 counts as infinite). Each run is then evaluated, with stochastic actions, on 256 episodes that were never used for training; all numbers in the tables come from this separate evaluation, not from training logs. There is no
wall-clock cap; a process that produces no epoch for 40 minutes (PhysX device hang) is
restarted from its latest checkpoint. Runs of the same species are interleaved so that every
species advances at the same rate. All metrics are logged to Weights \& Biases every epoch.

\paragraph{Fish selection.} Two fish per species are used for Task~1 and Task~2, and one per
species for the Case Study.

\section{Compute}
All experiments run on one workstation with two NVIDIA RTX PRO 6000 (96\,GB) GPUs. A single
trajectory-following run at 256 environments takes about 10\,min per PPO epoch when two runs
share a GPU; the whole box saturates at $\approx0.4$ epochs/min regardless of the number of
concurrent runs. Mesh generation costs about
30 Meshy credits per fish; the actor--critic loop takes 13--40\,min per fish on one GPU
(shared VLM).

\section{Manually Constructed Skeleton Templates}
\label{app:skeleton-dataset}

The actor is fine-tuned on, and the template library is derived from, fourteen fish models
whose skeletons were placed by hand. This section documents how they were made.

\paragraph{Species and meshes.}
The set covers fourteen body plans spanning the shapes in our video dataset and beyond: carp,
pumpkinseed, red snapper, roach, ruby, rudd, salmon, sea bass, spotted seatrout, tench, tiger
barb, tilapia, trevally and zander. Each starts from a closed, textured fish mesh in Blender
(metric units, length axis $+x$, head at $+x$, dorsoventral axis $y$), canonicalized exactly as
the pipeline meshes are (Sec.~\ref{app:pipeline}). Body length-to-height ratios range from
1.8 (tiger barb, pumpkinseed) to 3.8 (salmon).

\paragraph{Bone placement.}
Bones are axis-aligned box segments, placed one at a time along the longitudinal body midline
from the skull toward the caudal peduncle, following published descriptions of teleost axial
skeletons and body mechanics \citep{lauder2015fish,videler1993fish}. Three rules were enforced
while placing them: (i) the box center lies on the body midline in both the lateral and the
dorsal view; (ii) no box vertex may leave the skin (checked in Blender by intersecting each box
with the mesh); (iii) consecutive boxes abut so the chain is continuous from the first
post-cranial segment to the caudal peduncle. Segment count follows the visible flexibility of
the species: 4 (pumpkinseed) to 8 (zander), 85 bones in total, median 6. Boxes are not rotated
(all rotations are identity), so a bone is fully described by its center and size; the head
and the caudal fin themselves are left unarticulated, which is why the expert chains span only
38--64\% of body length (Sec.~\ref{app:skeleton}). No armature or skinning is authored: the
mesh--bone coupling is produced later by the FEM attachment step, identically for these
models and for the pipeline assets.

\paragraph{Joints and physical parameters.}
The hand-placed models carry no per-bone joint parameters. Joint type (3-DoF D6), limits
($\pm80^\circ$), PD gains ($k_p=240$, $k_d=6$) and the per-bone mass rule (body volume
distributed in proportion to bone volume) are a single physics preset applied by the export
step to every fish, so the annotator's decisions are limited to \emph{where} bones go and
\emph{how large} they are. The preset values were chosen from the ranges reported for fish
intervertebral joint mechanics and body stiffness \citep{jimenez2023flexibility}
and validated in simulation on the salmon model before being frozen.

\paragraph{From models to a template library and training data.}
Each model is exported to a \texttt{skeleton.json} (bone centers, sizes, world transforms and
the mesh bounding box). Two artifacts are built from these files:
\begin{itemize}\setlength{\itemsep}{1pt}
\item \emph{Template library.} The actor prompt lists every template by name with its bone
  count and body-shape ratios (length/height, length/thickness); the actor selects the template
  whose body plan best matches the target fish and edits its bones (add / remove / reposition /
  resize) inside the new mesh.
\item \emph{Fine-tuning set.} For every model the bones are replayed one at a time in
  head-to-tail order; at step $i$ the partial articulation (bones $0\ldots i{-}1$) is rendered
  in the three views used at inference, and the pair (renders, current articulation) $\to$
  ``add bone $i$ with this center and size'' is one supervised example. Roach and trevally are
  held out for validation, leaving 74 training examples from the other twelve models, which
  is the set the LoRA actor is trained on (Sec.~\ref{app:skeleton}).
\end{itemize}

\subsection{Hydrodynamics}
\label{app:hydro}
 
Resolving the flow around a deformable body at every physics step is far too
slow for the scale of training this benchmark requires, so we replace the fluid
solver with a resistive panel model: the force on each body element is taken to
be the force on a corresponding element of a straight cylinder moving at the
same speed and inclination \citep{taylor1952analysis}. The deformable skin is
partitioned into per-bone slices, and each surface panel $p$ of a slice (area
$A_p$, outward normal $\hat{\mathbf{n}}_p$) receives a quadratic drag force with
normal (form-drag) and tangential (skin-friction) components,
\begin{equation}
\begin{aligned}
    \mathbf{f}_p &= -\tfrac{1}{2}\rho A_p
    \left( C_n\,|v_{p,n}|\,v_{p,n}\,\hat{\mathbf{n}}_p
         + C_t\,\|\mathbf{v}_{p,t}\|\,\mathbf{v}_{p,t} \right),
    \\
    v_{p,n}&=\mathbf{v}_p\!\cdot\!\hat{\mathbf{n}}_p,\\
    \mathbf{v}_{p,t}&=\mathbf{v}_p - v_{p,n}\hat{\mathbf{n}}_p,
\end{aligned}
\end{equation}
where $\mathbf{v}_p$ is the panel velocity induced by the motion of its bone,
$\rho$ is the water density, and $C_n = 1.0$, $C_t = 0.01$ are the flat-plate
form-drag and skin-friction coefficients. Panel forces and their moments about
the bone origin are summed into one wrench per bone and applied at every physics
step.

\paragraph{Reactive force.}
Drag alone gives an undulating body little thrust, so we add the reactive force
of Lighthill's large-amplitude elongated-body theory \citep{lighthill1971large}.
Each bone $i$ carries the momentum of the water that its cross-section sets in
motion, $\mathbf{p}_i = m_{n,i}\,w_{n,i}\,\hat{\mathbf{n}}_i +
m_{b,i}\,w_{b,i}\,\hat{\mathbf{b}}_i$, where $w_{n,i}$ and $w_{b,i}$ are the
lateral and dorso-ventral velocities of the section. The added masses follow
from the depth $h_i$ and width $b_i$ of the section in the rest pose and the
bone length $L_i$: $m_{n,i}=C_a\rho\pi(h_i/2)^2L_i$ and
$m_{b,i}=C_a\rho\pi(b_i/2)^2L_i$, with the physical value $C_a=1$ for every
fish; no coefficient is fitted. This momentum is convected toward the tail at
the forward speed $u_i$ of the section, with the flux
\begin{equation}
    \boldsymbol{\Phi}_i = \frac{1}{L_i}\Big(u_i\,\mathbf{p}_i
    - \tfrac{1}{2}\big(m_{n,i}w_{n,i}^2+m_{b,i}w_{b,i}^2\big)\hat{\mathbf{t}}_i\Big),
\end{equation}
where $\hat{\mathbf{t}}_i$ points toward the head. The force on bone $i$ is the
flux that enters from its head-side neighbour minus $\boldsymbol{\Phi}_i$.
Summed over the body, these terms leave only the flux at the trailing edge,
whose average over a tail-beat cycle is Lighthill's thrust. The inertial part
of the reaction, $-\dot{\mathbf{p}}_i$, is included implicitly: the mean of
$m_{n,i}$ and $m_{b,i}$ is added to the mass of bone $i$ in the rigid-body
solver, which is more stable than an explicit force. Because the generated
meshes often lack a full caudal fin, the section depth of the last two bones is
at least $0.15$\,BL.

\paragraph{Numerical limits.}
The total hydrodynamic force and torque on each bone are clipped at 5\,N and
0.05\,N\,m, and the reactive force at $50\,\mathrm{BL/s^2}$ times the bone
mass. These limits only guard against numerical runaway.

The model does not resolve the flow and therefore neglects vortex shedding
and wake history, which a full fluid solver resolves
\citep{liu2022fishgym,yeo2010simulation}. Absolute swimming speeds should be read as
specific to this simulator rather than as predictions about the animal. What
the benchmark requires is weaker: that the same approximation apply to every
method and every fish, so that differences between methods are attributable to
the controller rather than to the fluid model.
\section{Human Study Details}
\label{app:human-study}

\paragraph{Clips.}
The study uses three fish with clean front-view video, one each from three
species: brook trout 019, channel catfish 002 and lake sturgeon 001. For these
fish we take five held-out 3\,s windows from the Task~1 test split: two from
brook trout, two from catfish and one from sturgeon. For each window and each
of the five Task~1 methods (Joint RL, Joint RL + AMP, CPG + RL, BCO + RL,
BLM + RL), we load the seed-0 checkpoint with the best training completion
rate. We then run 64 parallel episodes from the real fish's initial state,
using stochastic actions, and render one of them drawn at random. The same rule is applied
to every method. No best-of-64 selection is made, so each clip is a random sample of what
the method does on that window. This gives $5\times5=25$ method clips.

\paragraph{Anchors.}
We add two lower-anchor clips (catfish 002 and brook trout 019), in which the
fish receives zero actions for the whole window and starts with a small
backward velocity ($-0.1\times$ the real initial velocity). The anchor fish
does not swim, so it should receive low scores. These clips let us check that
raters do not reward a fish that does not swim.

\paragraph{Rendering.}
All clips are rendered in the simulator from a camera placed at the position
of the real front camera, and play at real-time speed in a loop. In each
window, all methods share one crop, the union of the bounding boxes of all
five methods, so that the crop does not reveal how far a fish travelled.
Part~1 videos show the simulated fish alone, on a plain background and
without the reference path. Part~2 videos show the real front-view recording
(left) and the simulated fish (right), side by side and starting at the same
time. Every clip gets a random two-character code, and the mapping from codes
to methods is never shown to raters. All raters see the clips in the same
order, sorted by code; because the codes are random, this order is unrelated
to the method or the fish.

\paragraph{Questions.}
Before the first clip, raters see a short introduction that explains the two
parts. It says that most clips are computer simulations and that a few are
checks. Raters then give their consent. Every question uses a 1--5 scale, with
the two ends labelled as shown in Table~\ref{tab:human-items}. Raters can go
back to any earlier clip and change their answers.

\begin{table}[H]
\caption{Human-study questions, exactly as shown to raters.}
\label{tab:human-items}
\begin{center}
\small
\begin{tabular}{@{}p{0.12\linewidth}p{0.62\linewidth}p{0.2\linewidth}@{}}
\toprule
Item & Question & Scale (1 / 5) \\
\midrule
\multicolumn{3}{@{}l}{\emph{Part 1: the test fish alone}} \\
Plausible & How possible is it for a real fish to make this movement? & not possible / fully possible \\
\midrule
\multicolumn{3}{@{}l}{\emph{Part 2: real fish (left) and test fish (right), same path, same start time}} \\
Path & Does the test fish (right) swim the same route as the real fish (left), including going up or down? \emph{Please ignore differences in body pose (how the body bends or tilts); compare only the route.} & very different / the same \\
Speed match & Can the test fish swim as fast as the real fish? \emph{The simulated fish starts from rest and needs a short time to speed up, while the real fish is already swimming. Please ignore this speed-up and when each fish arrives; judge only its top swimming speed.} & much slower / as fast \\
Appearance & Does the test fish's body look like the real fish (shape, size, fins, colour and texture)? Ignore how it moves. & very different / the same \\
Similarity & Taking everything into account, how similar is the test fish to the real fish? & very different / the same \\
\midrule
\multicolumn{3}{@{}l}{\emph{After the last clip}} \\
Confidence & How confident are you in your ratings? & not confident at all (mostly guessing) / very confident \\
\bottomrule
\end{tabular}
\end{center}
\end{table}

\paragraph{Raters and procedure.}
Raters are twelve people who work with fish professionally, recruited from
fish hatcheries. The survey is a web page
that requires no account. Answers are stored without the rater's name. At the
end, the page shows a short result code that holds the answers, and the rater
emails this code back to us. The study takes about 15 minutes. For each clip,
we also record the time spent on Part~1 and Part~2 as an attention check.

\paragraph{Analysis.}
Table~\ref{tab:human} reports the raw 1--5 ratings. For each method and each item, we pool
the ratings of its five clips from all twelve raters (60 ratings; 24 for the two anchor clips)
and report their mean and standard deviation. No rater was excluded.

\section{Novelty Compared with Prior Work}
\label{app:novelty}
Table~\ref{tab:novelty} compares Video2SwimFish item by item with the closest prior work
discussed in Section~\ref{sec:related}.

\begin{table}[H]
\caption{What is new in Video2SwimFish, compared item by item with the closest prior work.
Numbers are from Table~\ref{tab:task1} and Section~\ref{sec:mesh-reconstruction}.}
\label{tab:novelty}
\centering
\footnotesize
\setlength{\tabcolsep}{3pt}
\renewcommand{\arraystretch}{1.12}
\begin{tabular}{@{}p{0.03\linewidth}p{0.28\linewidth}p{0.30\linewidth}p{0.34\linewidth}@{}}
\toprule
 & \textbf{Prior work} & \textbf{Video2SwimFish} & \textbf{Difference and value} \tabularnewline
\midrule
\multicolumn{4}{@{}l}{\emph{Asset}} \tabularnewline
1 & \raggedright Dynamic NeRF, BANMo \citep{pumarola2021dnerf,park2021nerfies,yang2022banmo}: deforming surface and appearance from video; no bones, joints or metric scale
  & \raggedright Hybrid asset at real size: a rigid articulation placed inside a soft FEM body
  & \raggedright Both deformable and controllable: enables physics simulation and RL of a deforming fish \tabularnewline
2 & \raggedright Single-image 3D \citep{hong2024lrm,tochilkin2024triposr,xu2024instantmesh}: static mesh, arbitrary scale, any object
  & \raggedright Image-to-3D is one step; a VLM selects the canonical frame and the calibrated views give the scale
  & \raggedright Each mesh is one recorded individual, checked against outlines of the real fish (IoU 0.91 side view, 0.75 unseen top view) \tabularnewline
3 & \raggedright URDFormer, Articulate-Anything \citep{chen2024urdformer,le2024articulate}: articulation of rigid objects with separate parts
  & \raggedright VLM actor--critic places a bone chain inside one continuous deformable body
  & \raggedright Automatic articulation extends from rigid objects to soft animals; 120 fish are rigged automatically \tabularnewline
\midrule
\multicolumn{4}{@{}l}{\emph{Swimming policy}} \tabularnewline
4 & \raggedright Imitation from observation \citep{torabi2018bco,edwards2019ilpo}: actions inferred by an inverse dynamics model
  & \raggedright Video midline mapped to joint space by inverse kinematics; no action labels
  & \raggedright No pseudo-label error from a learned dynamics model; motion coverage 0.85 against 0.78 for BCO+RL \tabularnewline
5 & \raggedright DeepMimic, AMP \citep{peng2018deepmimic,peng2021amp}: joint-level reference motion used as a reward
  & \raggedright Real motion is encoded as the low-dimensional action space (BLM), not as a reward term
  & \raggedright No motion capture needed; faster training; completion rate 0.70 against 0.46 for Joint RL+AMP \tabularnewline
6 & \raggedright CPG for fish robots \citep{ijspeert2008central,xie2019central}: hand-designed rhythm, shared by all individuals
  & \raggedright One manifold per fish, extracted from its own 3D motion
  & \raggedright The motion prior is individual and comes from data; motion coverage 0.85 against 0.74 for CPG+RL \tabularnewline
\midrule
\multicolumn{4}{@{}l}{\emph{Dataset and benchmark}} \tabularnewline
7 & \raggedright 3D-ZeF \citep{pedersen20203d}: several fish of one species per sequence; 3D trajectories
  & \raggedright 120 individuals of 6 species, one fish per synchronized two-view recording, plus 120 simulated fish with policies
  & \raggedright Scale and scalability; moving targets for underwater robot learning; each simulated fish can be evaluated against its own real fish \tabularnewline
8 & \raggedright FishGym \citep{liu2022fishgym} and embodied benchmarks \citep{gu2023maniskill2,li2024behavior1k}: hand-designed assets, scored by task success
  & \raggedright Task success and fidelity to the individual (path, body bending, speed), an expert human study and a robot case study
  & \raggedright Shows that task success and biological fidelity do not always improve together, and gives metrics to track both \tabularnewline
\bottomrule
\end{tabular}
\end{table}

\subsection*{Ethics statement}

\textbf{Animal subjects.} This work involves video recording of live
fish. All procedures were reviewed and approved by our institution's
Institutional Animal Care and Use Committee (IACUC) prior to data
collection, under an approved protocol covering observational
video-based study of fish for aquaculture research
(Protocol No.\ LA26-0016).
Our study is purely observational: fish were recorded while swimming
freely in a holding tank, and no invasive procedure, restraint, tagging,
or experimental manipulation was performed at any point. No animals were
euthanized for this study. Recording sessions were kept short, water
quality and lighting were controlled throughout, and animals were
returned to their holding conditions immediately after each session. We
note that reducing reliance on live-animal experimentation is itself
part of the motivation for this work: the resulting simulated fish are
intended to let embodied agents be trained without repeated interaction
with real animals.

\filbreak
\textbf{Dataset release.} We release both the multi-view fish video
dataset and the derived simulation assets at
\url{https://huggingface.co/datasets/video2swimfish/video2swimfish-dataset}.
The recordings contain no human subjects and no personally identifiable
information. The datasets are released under the CC BY-NC 4.0 license for
non-commercial research use. The data collection protocol is documented
in Section~\ref{sec:data-collection} so that the conditions under which
the data were obtained are transparent to users; anyone reproducing our
collection procedure with live animals should obtain their own
institutional animal-ethics approval.

\textbf{Potential impact.} The intended application is training embodied
agents for monitoring and management tasks in aquaculture. We note that
agents trained to interact with or capture fish could, if deployed
without care, be used in ways that adversely affect animal welfare, and
we encourage downstream users to evaluate welfare implications before
real-world deployment.

\subsection*{Reproducibility statement}
We release the synchronized multi-view videos of all 120 fish and the 120
derived simulation assets, each with its mesh, articulation and trained
swimming policy, at
\url{https://huggingface.co/datasets/video2swimfish/video2swimfish-dataset}.
Section~\ref{sec:data-collection} and Appendix~\ref{app:dataset} describe the recording setup, including
cameras, tank dimensions and calibration, and give per-species details, so that new videos can be collected under the
same conditions. Section~3 describes each stage of the pipeline, and
Appendix~\ref{app:skeleton-dataset} describes the skeleton templates and the
hand-built fish models used to fine-tune the actor VLM. The pipeline relies
on an open-weight VLM, Qwen3-VL~\citep{qwen3vl2025}, and a commercial
image-to-3D service, Meshy~\citep{meshy_image_to_3d}, whose model may change
over time. Because of this, and because the actor--critic loop samples
stochastically, re-running the pipeline may not reproduce identical assets;
all benchmark results are therefore computed on, and reproducible from, the
released assets. Section~\ref{sec:setup} gives the simulation settings
(actuator preset, physics and control rates, hydrodynamic model and
coefficients), the BLM controller and the training and stopping criteria,
with the BLM decoder detailed in Appendix~\ref{app:sim}.
Sections~\ref{sec:task1}--\ref{sec:task3} define each task and metric, and Appendix~\ref{app:behavior} describes how the 3D midline and the joint-angle BLM are extracted from the two views. Source code for the pipeline, simulation environment,
benchmark tasks and baselines can be found at
\url{https://github.com/hangongchen/Video2SwimFish}.

\end{document}